\documentclass[10pt,journal,compsoc]{IEEEtran}
\usepackage{graphicx}
\usepackage{amsmath}
\usepackage{amssymb}
\usepackage{booktabs}
\usepackage{multirow}
\usepackage{hhline}
\usepackage{bbding} 
\usepackage[breaklinks,colorlinks]{hyperref}
\usepackage[accsupp]{axessibility}

\usepackage{makecell}
\usepackage[table,xcdraw]{xcolor}
\usepackage{threeparttable}

\usepackage{enumitem}

\usepackage{xcolor}

\usepackage{booktabs}

\usepackage{caption}

\ifCLASSOPTIONcompsoc
  \usepackage[nocompress]{cite}
\else
  \usepackage{cite}
\fi
\ifCLASSINFOpdf
\else
\fi
\begin{document}
%

\title{A Survey on Facial Expression Recognition of Static and Dynamic Emotions}
%
%

%
%

\author{Yan~Wang, Shaoqi~Yan, Yang~Liu, Wei~Song, Jing~Liu, Yang~Chang, Xinji~Mai,  
\protect\\ 
Xiping~Hu, Wenqiang~Zhang* and Zhongxue~Gan* 
        
        
        
\IEEEcompsocitemizethanks{\IEEEcompsocthanksitem Yan~Wang, Shaoqi~Yan, Yang~Liu, Jing~Liu, Yang~Chang and Xinji~Mai are with the Academy for Engineering \& Technology, Fudan University, Shanghai, China. E-mail:  \{yanwang19, sqyan19, yang\_liu20, jingliu19\}@fudan.edu.cn; {ychang24, xjmai23}@m.fudan.edu.cn.


\IEEEcompsocthanksitem Wei Song is with the College of Information Technology, Shanghai Ocean University, Shanghai, China. E-mail: wsong@shou.edu.cn.

\IEEEcompsocthanksitem Xiping Hu is with the School of Medical Technology, Beijing Institute of Technology, Beijing, China. E-mail: huxp@bit.edu.cn.

\IEEEcompsocthanksitem Wenqiang Zhang is with the Academy for Engineering \& Technology, Fudan University, Shanghai, China, and also with the School of Computer Science, Fudan University, Shanghai, China. E-mail: wqzhang@fudan.edu.cn.

\IEEEcompsocthanksitem Zhongxue Gan is with the Academy for Engineering \& Technology, Fudan University, Shanghai, China. E-mail: ganzhongxue@fudan.edu.cn.

}
\thanks{Manuscript received August 28, 2024;\\(*Corresponding authors: Zhongxue Gan and Wenqiang Zhang.)}}

%
%

\markboth{IEEE TRANSACTIONS ON PATTERN ANALYSIS AND MACHINE INTELLIGENCE, August~2024}%
{Shell \MakeLowercase{\textit{Yan Wang et al.}}: Bare Demo of IEEEtran.cls for Computer Society Journals}
%




\IEEEtitleabstractindextext{
\begin{abstract}

Facial expression recognition (FER) aims to analyze emotional states from static images and dynamic sequences, which is pivotal in enhancing anthropomorphic communication among humans, robots, and digital avatars by leveraging AI technologies. As the FER field evolves from controlled laboratory environments to more complex in-the-wild scenarios,  advanced methods have been rapidly developed and new challenges and apporaches are encounted, which are not well addressed in existing reviews of FER. This paper offers a comprehensive survey of both image-based static FER (SFER) and video-based dynamic FER (DFER) methods, analyzing from model-oriented development to challenge-focused categorization. We begin with a critical comparison of recent reviews, an introduction to common datasets and evaluation criteria, and an in-depth workflow on FER to establish a robust research foundation. We then systematically review representative approaches addressing eight main challenges in SFER (such as expression disturbance, uncertainties, compound emotions, and cross-domain inconsistency) as well as seven main challenges in DFER (such as key frame sampling, expression intensity variations, and cross-modal alignment). Additionally, we analyze recent advancements, benchmark performances, major applications, and ethical considerations. Finally, we propose five promising future directions and development trends to guide ongoing research. The project page for this paper can be found at \url{https://github.com/wangyanckxx/SurveyFER}.

\end{abstract}

}

\maketitle

\IEEEdisplaynontitleabstractindextext

%
\IEEEpeerreviewmaketitle

\section{Introduction \label{sec:introduction}}


%
%
%
%

\IEEEPARstart{A}{FFECVIVE} computing~\cite{Zhao_10253654} has far-reaching influence and importance in key national fields. Innovate UK, the UK's innovation agency, identified "artificial intelligence (AI) emotion and expression recognition" as the top among 50 emerging technologies\footnote{ \url{https://www.ukri.org/wp-content/uploads/2023/12/IUK-05122023-INO0617_Emerging-Tech-Report_AW2-final.pdf}} that would have profoundly influence the British economy and society in 2024. The China Association for Science and Technology grandly released the major scientific issues of 2024, among which the research on digital humans and robots with emotions and emotional intelligence was selected as one of the top ten frontier scientific issues\footnote{\url{https://www.cast.org.cn/xw/KXYW/art/2024/art_e9df73f3c2f5480aaaa6e78fffd69acd.html}}. Clearly, the development of AI emotion and expression recognition technology has become an inevitable requirement for general AI, digital computing and multi-disciplinary research~\cite{wang2023unlocking}.

Facial expressions~\cite{krumhuber2023role} are the primary and straightforward means of human emotional expression, frequently employed and of utmost importance in interpersonal interaction~\cite{cowen2021sixteen, WangJZ_10205489}. They convey richer affective information non-verbally than other forms of messages like voice, gestures, and body postures~\cite{ wang2022systematic}. The concept of facial emotions was originally introduced by Darwin in his book "The Expression of the Emotions in Man and Animals" (1872). It has been noted that expressions are innate in nature and the remains of adaptive movements of animals and humans during evolution and survival. Ekman and Friesen \cite{ekman1979facial} proposed six basic emotions: Happy, Angry, Sad, Surprise, Fear, and Disgust, and found universal associations between specific facial muscle patterns and emotions types, which are consistent across cultures. In recent years, with the advancement of AI technologies, facial emotion recognition (FER) methods have rapidly developed and shown wide applications in psychological research~\cite{rudovic2018personalized}, medical diagnosis~\cite{savchenko2022classifying}, and intelligent human-computer interaction~\cite{Zhaosc_7744512_2018}.


\begin{figure*}[ht]
\vspace{-0.1cm}
  \centering
   \includegraphics[width=1.0\linewidth]{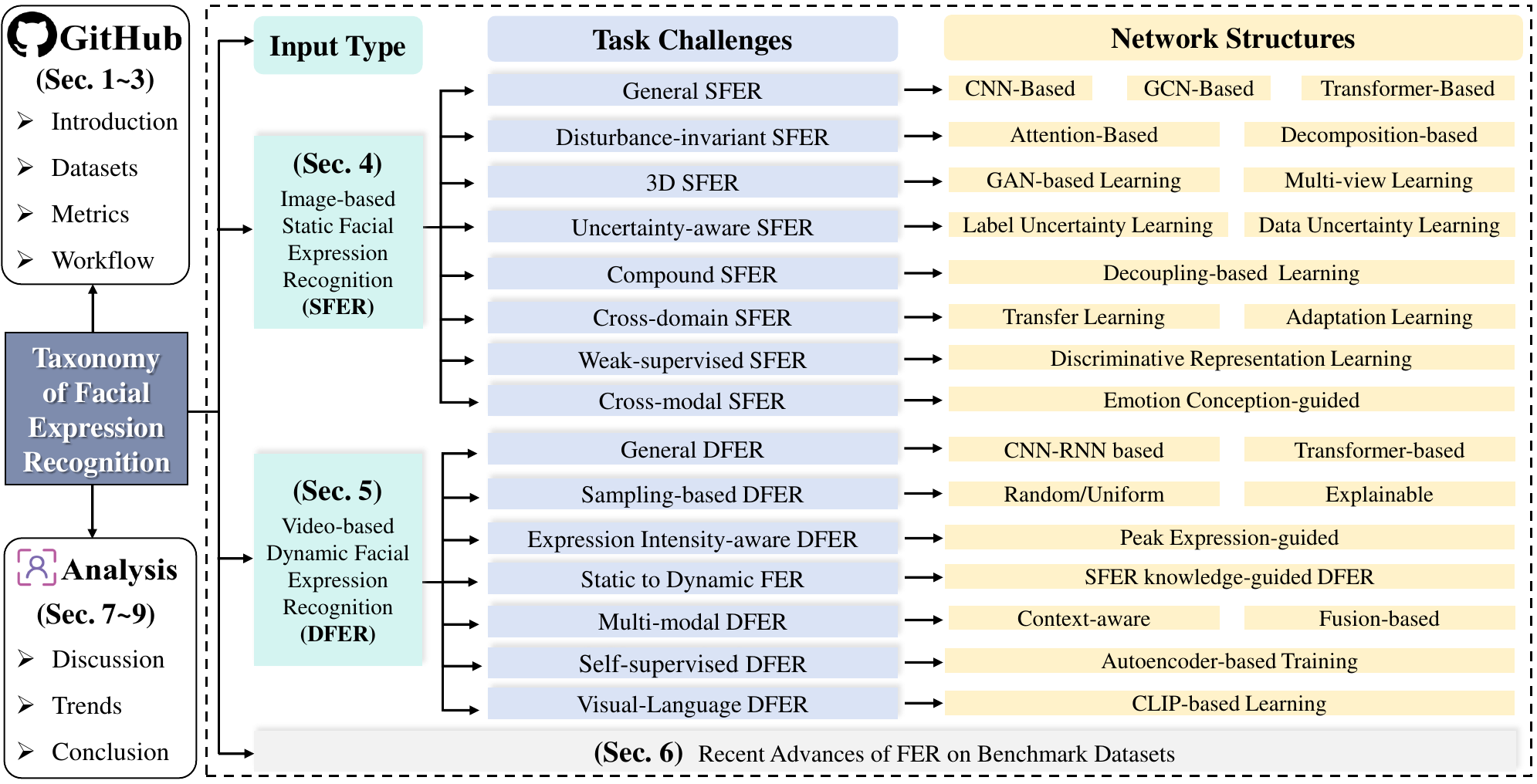}
   \vspace{-0.5cm}
   \caption{Taxonomy of FER of static and dynamic emotions. We present a hierarchical taxonomy that categorizes existing FER models by input type, task challenges, and network structures within a systematic framework, aiming to provide a comprehensive overview of the current FER research landscape. First, we have introduced datasets, metrics, and workflow (including literature and codes) into a public GitHub repository\protect\footnotemark ~\textbf{(Sec. \ref{sec:introduction}, \ref{sec:MULTI_SCENE}, and \ref{sec:Tutorial}}). Then, image-based SFER \textbf{(Sec. \ref{sec:Static})} and video-based DFER \textbf{(Sec. \ref{sec:Dynamic})} overcome different task challenges using various learning strategies and model designs. Following, we analyzed recent advances of FER on benchmark datasets \textbf{(Sec. \ref{sec:Discussion})}. Finally, we discuss and conclude some important issues and potential trends in FER, highlighting directions for future developments \textbf{(Sec. \ref{sec:Applications}, \ref{sec:Development}, and \ref{sec:Conclusion}}).}
   \label{fig:taxonomy}
   \vspace{-0.3cm}
\end{figure*}

The FER aims to identify an individual's emotional state based on the analysis of facial expressions~\cite{wang2019novel, Liu_10073607}. Depending on the type of data used to capture the expressions, the FER can be divided into two parts: image-based static FER (SFER)~\cite{CEPrompt_zff_2024, wang2024mgr} and video-based dynamic FER (DFER)~\cite{jiang2020dfew, wang2022dpcnet, mai2024all}. The SFER works on solving challenges due to pose occlusion, cross-domain inconsistency, label uncertainty, insufficient data volume, and cross-modality. Researchers also use various data augmentation techniques and regularization methods to alleviate the problems of insufficient data volume and label uncertainty. In addition, the robustness and accuracy of expression recognition are enhanced through cross-modal information fusion. While SFER focuses on instantaneous expressions, DFER concentrates on temporal changes of facial expressions to accurately describe and comprehend the whole process of emotional shifts. Dealing with expression recognition in video sequences, DFER has main challenges in key frame extraction, spatiotemporal feature extraction, expression intensity changes, and cross-modal fusion. To capture the dynamic expression information, DFER models not only focus on static features in a single frame, but also incorporate the temporal relationship between consecutive frames.

\begin{table*}[]
\centering
\caption{Comparisons on our FER review with state-of-the-art FER-related reviews from 2020 to 2024.}
\vspace{-0.2cm}
\resizebox{\textwidth}{!}{
\setlength\tabcolsep{4pt}
\begin{tabular}{@{}lccccccccccc|cccccc|ccc@{}}
\toprule
\multicolumn{1}{c}{\multirow{2}{*}{\textbf{Pub. {[}Ref{]}}}} & \multirow{2}{*}{\textbf{Year}} & \multicolumn{2}{c}{\textbf{Datasets}} & \multirow{2}{*}{\textbf{WF}} & \multicolumn{7}{c|}{\textbf{Image-based Static FER}} & \multicolumn{6}{c|}{\textbf{Video-based Dynamic FER}} & \multicolumn{3}{c}{\textbf{Application}} \\ \cmidrule(lr){3-4} \cmidrule(l){6-21} 
\multicolumn{1}{c}{}& & S & D  && DI& 3D& UA&  CP&  CD&  LS &  CM&  SL&  EI&  MM&  SD&  SS& VL  & HPC & PE & HCI\\ \midrule \midrule
IEEE TAFFC {\cite{li2020deep}}   & 2022   &\Checkmark  & \Checkmark  & \Checkmark  & \Checkmark & \Checkmark & \Checkmark & \XSolidBrush & \XSolidBrush  & \XSolidBrush  & \XSolidBrush  & \XSolidBrush  & \Checkmark  & \XSolidBrush   & \XSolidBrush   & \XSolidBrush  & \XSolidBrush & \XSolidBrush & \XSolidBrush   & \Checkmark   \\
INFSCI {\cite{canal2022survey}}  & 2022   &\Checkmark  & \XSolidBrush  & \XSolidBrush  & \XSolidBrush & \XSolidBrush & \XSolidBrush & \XSolidBrush & \XSolidBrush  & \XSolidBrush  & \XSolidBrush  & \XSolidBrush  & \XSolidBrush  & \XSolidBrush   & \XSolidBrush   & \XSolidBrush  & \XSolidBrush & \XSolidBrush & \XSolidBrush   & \XSolidBrush    \\
IEEE TIE {\cite{10041168_survey_2023}}  & 2023   &\Checkmark  & \Checkmark  & \Checkmark  & \Checkmark & \Checkmark & \Checkmark & \Checkmark & \Checkmark  & \XSolidBrush  & \XSolidBrush  & \XSolidBrush  & \Checkmark  & \XSolidBrush   & \XSolidBrush   & \XSolidBrush  & \XSolidBrush & \Checkmark & \XSolidBrush   & \Checkmark   \\
COMSCIREV  {\cite{leong2023facial}}& 2023   & \XSolidBrush  & \XSolidBrush  & \Checkmark  & \XSolidBrush & \XSolidBrush & \Checkmark & \XSolidBrush & \Checkmark  & \XSolidBrush  & \XSolidBrush  & \XSolidBrush  & \XSolidBrush  & \XSolidBrush   & \XSolidBrush   & \XSolidBrush  & \XSolidBrush & \XSolidBrush & \XSolidBrush   & \Checkmark   \\
PR {\cite{alexandre2020systematic}}  & 2020   & \Checkmark  & \Checkmark  & \Checkmark  & \XSolidBrush & \Checkmark & \XSolidBrush & \Checkmark & \Checkmark  & \XSolidBrush  & \XSolidBrush  & \XSolidBrush  & \XSolidBrush  & \XSolidBrush   & \XSolidBrush   & \XSolidBrush & \XSolidBrush  & \XSolidBrush & \XSolidBrush   & \XSolidBrush   \\
IEEE TAFFC {\cite{9802683_Survey}}& 2022   & \XSolidBrush  & \XSolidBrush  & \Checkmark  & \XSolidBrush & \XSolidBrush & \Checkmark & \XSolidBrush & \Checkmark  & \XSolidBrush  & \XSolidBrush  & \XSolidBrush  & \XSolidBrush  & \XSolidBrush   & \XSolidBrush   & \XSolidBrush  & \XSolidBrush & \XSolidBrush & \Checkmark   & \XSolidBrush   \\
IEEE TAFFC {\cite{Zhaogy9924612}}  & 2023   & \Checkmark  & \Checkmark  & \Checkmark  & \XSolidBrush & \XSolidBrush & \Checkmark & \Checkmark & \Checkmark  & \XSolidBrush  & \XSolidBrush  & \XSolidBrush  & \XSolidBrush  & \XSolidBrush   & \XSolidBrush   & \XSolidBrush & \XSolidBrush  & \XSolidBrush & \XSolidBrush   & \XSolidBrush   \\
IEEE TPAMI {\cite{10420507_ddb_2024}}  & 2023   & \Checkmark  & \XSolidBrush  & \XSolidBrush  & \XSolidBrush & \XSolidBrush & \XSolidBrush & \XSolidBrush & \Checkmark  & \XSolidBrush  & \XSolidBrush  & \XSolidBrush  & \XSolidBrush  & \XSolidBrush   & \XSolidBrush   & \XSolidBrush & \XSolidBrush  & \Checkmark & \Checkmark   & \Checkmark   \\
IEEE TPAMI {\cite{hassan2019automatic}} & 2021   & \Checkmark  & \Checkmark  & \Checkmark  & \Checkmark & \XSolidBrush & \XSolidBrush & \XSolidBrush & \XSolidBrush  & \XSolidBrush  & \XSolidBrush  & \XSolidBrush  & \XSolidBrush  & \XSolidBrush   & \XSolidBrush   & \XSolidBrush & \XSolidBrush  & \Checkmark & \Checkmark   & \Checkmark   \\
\midrule
Our FER Review & 2024   & \Checkmark  & \Checkmark  & \Checkmark  & \Checkmark & \Checkmark & \Checkmark & \Checkmark & \Checkmark  & \Checkmark  & \Checkmark  & \Checkmark  & \Checkmark  & \Checkmark   & \Checkmark   & \Checkmark & \Checkmark  & \Checkmark & \Checkmark   & \Checkmark   \\ \bottomrule
\end{tabular}
}
\begin{tablenotes}
  \item \Checkmark and \XSolidBrush denote the corresponding content contains and does not contain systematic analysis and discussion, respectively. 
  \item S, D and WF represents Static, Dynamic and Workflow, respectively.
  \item DI, 3D, UA, CP, CD, LS and CM represents Static, Dynamic and Workflow, respectively. SL, EI, MM, SD, SS, and VL represents Sampling, Expression Intensity, Multi-modal, Static to Dynamic, Semi-supervised and Visual-Language, respectively. 
  \item HPC, PE, and HCI represents Health and Psychological Counseling, Personalized Education, and Human-Computer Interaction, respectively. 

  \end{tablenotes}
  \vspace{-0.5cm}
\label{tab:ComparisonFERReview}
\end{table*}

\footnotetext{\url{https://github.com/wangyanckxx/SurveyFER}}

\subsection{Taxonomy Overview}
\begin{figure}[t]
\vspace{-0.1cm}
  \centering
   \includegraphics[width=1.0\linewidth]{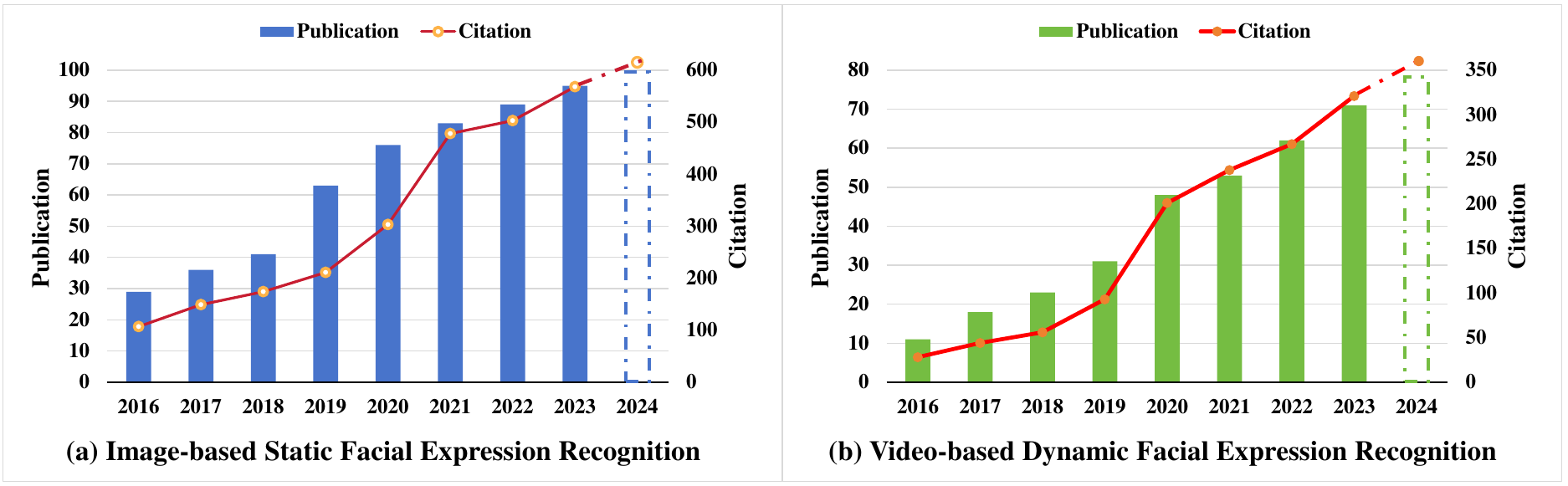}
   \vspace{-0.3cm}
   \caption{The statistics of Publication (Bar) and Citation (Line) on the topic of (a) image-based SFER and (b) video-based DFER from 2016 to 2024.}
   \label{fig:static}
   \vspace{-0.5cm}
\end{figure}

In this paper, we systematically summarize the current state of FER research and provide a hierarchical taxonomy to organize existing FER works according to input type (image-based SFER and video-based DFER), task challenges, and network structures, as shown in Fig.~\ref{fig:taxonomy}.  
For SFER, we identify eight key challenges such as disturbances, uncertainty, compound labels, cross-domain adaptability, and cross-modality issues, and summarize the model structures of existing approaches that often used to address the corresponding challenge. For DFER, we incorporate seven additional considerations like key frame extraction, expression intensity changes, static-to-dynamic consistency, semi-supervised learning, and cross-domain alignment, as well as the solutions of current methods. We further analyzed and discussed the recent advances of typical reviewed methods on benchmarking datasets. In addition, we have summarized the benchmark datasets, evaluation metrics, literature, codes, workflow, and discussions in the github repositories. To develop this taxonomy, we have extensively reviewed a substantial amount of research papers from 2016 to 2024. Fig.~\ref{fig:static} tracks the publication and citation trends related to image-based SFER and video-based DFER from 2016 to 2024. There is a notable surge in both publications and citations starting around 2019, continuing to rise through 2023 and projected into 2024. This reflects growing interest and advancements in both SFER and DFER fields.

\vspace{-0.1cm}





\subsection{Related Reviews}

In the past five years, some reviews~\cite{leong2023facial, Zhaogy9924612, 9802683_Survey, li2020deep} have covered FER works and generated various classification systems. To highlight the unique contributions of our review, we compared with several existing key reviews and summarized it in Table~\ref{tab:ComparisonFERReview}. Review studies~\cite{li2020deep, canal2022survey, 10041168_survey_2023} mainly introduce and analyze various DL-based FER techniques from the laboratory-controlled environment to in-the-wild circumstances. Our review not only provides an in-depth analysis of the standard processes of FER systems and the challenges in practical deployments, such as pose occlusion, cross-domain inconsistency, and label uncertainty, but also discusses different methods and technological advancements in image-based SFER and video-based DFER, providing a more comprehensive perspective. ~\cite{leong2023facial} primarily focuses on the most popular techniques and current trends in visual emotion recognition. However, our work further refines the study of a single modality (facial expressions), delving into the latest methods and technical challenges of static and dynamic FER. Recent FER review works~\cite{alexandre2020systematic, Zhaogy9924612, 9802683_Survey, 10420507_ddb_2024} focus on 3D FER, graph-based or multi-view facial expression analysis, and discusses demographic biases in FER datasets. Our review covers a wider variety of deep learning techniques, not limited to graph methods and multi-view issues but further covers the latest advancements in cross-domain learning, cross-modal fusion and self-supervised learning.~\cite{hassan2019automatic} mainly focuses on the application of pain detection or emotional mimicry in educational settings through facial expressions.  Our review covers more application scenarios and potentials of FER in different fields, while discussing technical and ethical issues.

\subsection{Contribution Summary}

To clarify FER development and inspire future research, this survey covers research background, datasets, generic workflow, task challenges, methods, performance evaluation,  applications, ethical issues, and development trends. In summary, the main contributions of this work are fourfold:

\setlist[itemize]{leftmargin=0.5cm}
\begin{enumerate}

 \item
To the best of our knowledge, this is the first comprehensive survey that divides FER research into image-based SFER and video-based DFER, extending from model-oriented development to challenge-oriented taxonomy, and provides an in-depth analysis of the real-world challenges and solutions. 
  \item
We systematically review the latest representative methods of SFER regarding eight main challenges (such as expression disturbance, uncertainty, cross-domain inconsistency) and DFER regarding seven main challenges (key frame extraction, expression intensity variations, and cross-modal alignment).
  \item
We summarize, analyze and discuss recent advances and technical challenges of FER on diverse benchmark datasets under the setups of in-the-lab FER, in-the-wild SFER, and in-the-wild DFER. 
\item
This survey summarizes three field applications and ethical issues, and discuss development trends (such as zero-shot FER and embodied facial expression generation), aiming to provide a new perspective and guidance on FER systems.
\end{enumerate}

\definecolor{sfer}{RGB}{182,199,234}
\definecolor{dfer}{RGB}{200,229,179}

    

\begin{figure*}[ht]
\vspace{-0.1cm}
  \centering
   \includegraphics[width=1.0\linewidth]{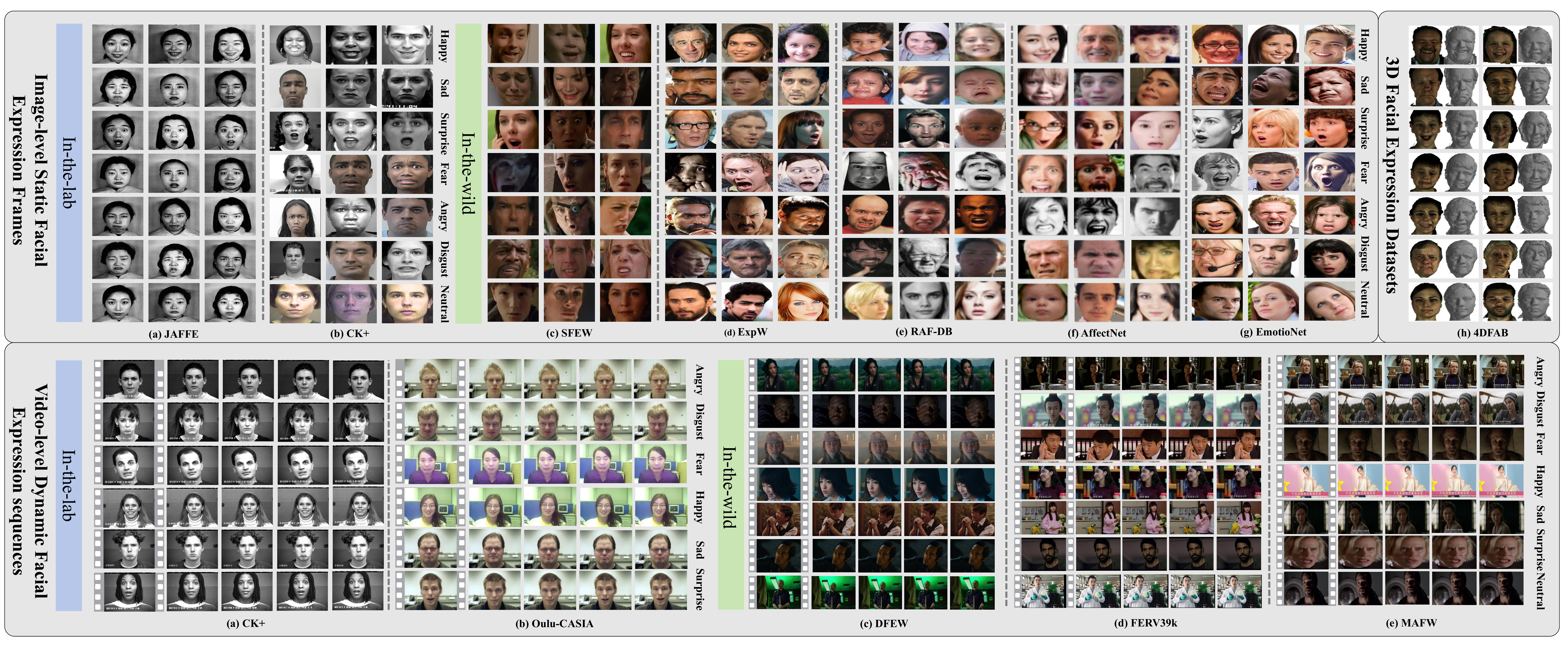}
   \vspace{-0.6cm}
   \caption{Image-based static facial frames (\textbf{Above}): (a) JAFFE~\cite{lyons1998coding}, (b) CK+~\cite{lucey2010extended}, (c) SFEW~\cite{dhall2011static}, (d) ExpW~\cite{zhang2018facial}, (e) RAF-DB~\cite{li2017reliable}, (f) AffectNet~\cite{mollahosseini2017affectnet}, (g) EmotioNet~\cite{fabian2016emotionet}, (h) 4DFAB~\cite{cheng20184dfab}; and video-based dynamic facial sequences (\textbf{Below}): (a) CK+~\cite{lucey2010extended}, (b) Oulu-CASIA~\cite{zhao2011facial}, (c) DFEW~\cite{jiang2020dfew}, (d) FERV39k~\cite{wang2022ferv39k}, and (e) MAFW~\cite{liu2022mafw} of seven basic emotions in the lab and wild.}
   \label{fig:Dataset}
   \vspace{-0.5cm}
\end{figure*}



\section{Datasets and Evaluation Metrics}
\label{sec:MULTI_SCENE}
Facial expression data is the key foundation for implementing and developing FER algorithms. Adequate and diverse expression datasets provide the necessary training and testing material for the FER algorithms. Table~\ref{tab:summary_datasets} shows the publicly available benchmark datasets with different attributes. Some examples of widely-used datasets are illustrated in Fig.~\ref{fig:Dataset}. Additionally, we introduce evaluation metrics.


\subsection{Image-based SFER Datasets}
The image-based SFER dataset is composed of individual images, each representing a specific emotional state. Based on the image collection scenarios, these datasets are divided into three groups: in-the-lab, in-the-wild, and 3D datasets .

For in-the-lab SFER, there are 5 widely used as benchmark datasets, including 1) \textbf{JAFFE}~\cite{lyons1998coding} consists of 213 images of seven basic facial expressions posed by 10 Japanese women, with each expression performed multiple times; 2) \textbf{CK+} ~\cite{lucey2010extended} includes 593 sequences, with 327 labeled for seven basic emotions plus contempt; expression intensity progresses from neutral to peak; 3) \textbf{Oulu-CASIA}~\cite{zhao2011facial} contains videos from 80 subjects under different lighting conditions, capturing six basic emotions; 4) \textbf{MMI}~\cite{valstar2010induced} includes 740 images and 2,900 video sequences depicting seven basic emotions, starting and ending with neutral expressions; 5) \textbf{RaFD}~\cite{langner2010presentation} features 8,040 high-quality images of seven basic facial expressions and contempt, taken from different angles with uniform settings, involving 67 professional actors.

For in-the-wild SFER, there are 6 widely used as benchmark datasets, including 1) \textbf{FER-2013}~\cite{goodfellow2013challenges} includes 35,887 images labeled with seven basic facial expressions; 2) \textbf{SFEW 2.0}~\cite{dhall2011static}, derived from AFEW, comprises 1,766 images; 3) \textbf{EmotioNet}~\cite{fabian2016emotionet} features over 1 million images labeled with basic and complex facial expressions, and action units (AUs); 4) \textbf{RAF-DB}~\cite{li2017reliable} contains 29,672 images, with 15,339 labeled for basic expressions and part compound expressions; 5)\textbf{AffectNet} ~\cite{mollahosseini2017affectnet}, collected using 1,250 emotion-related keywords in six languages, consists of approximately 1 million images, with seven discrete facial expressions and the intensity of valence and arousal; 6) \textbf{ExpW}~\cite{zhang2018facial} contains 91,793 images labeled with seven basic facial expressions.

For 3D/multi-view SFER, there are 4 widely used as benchmark datasets, including 1) \textbf{BU-3DFE}~\cite{yin20063d}, with 2,500 3D facial expressions from 100 subjects; 2) \textbf{Bosphorus} ~\cite{savran2008bosphorus} includes 4,652 3D facial images from 105 subjects, with a wide range of AUs and expressions; 3) \textbf{4DFAB}~\cite{cheng20184dfab} spans five years of collection from 180 subjects, comprising over 1.8 million high-resolution 3D facial images, including both posed and spontaneous expressions.

\begin{table}[!t]
    \centering
\caption{Summary of the in-the-lab or in-the-wild \textbf{{\color{magenta} datasets}} with static and dynamic emotions for FER training and evaluation. ECT: Elicitation; P: Posed; I: Instinctive; Sev: Seven Emotions (Happy, Angry, Surprise, Fear, Sad, Disgust, Neutral); C: Contempt; A: Anxiety;  D: Disappointment; H: Helplessness; Com: Compound.}
\renewcommand\arraystretch{1.4} 
\setlength{\tabcolsep}{0.4pt} 
\scalebox{0.85}{
    \begin{tabular}{
        >{\centering\arraybackslash}p{1.3cm} 
        >{\centering\arraybackslash}p{0.9cm} 
        >{\raggedright\arraybackslash}p{2.2cm} 
        >{\centering\arraybackslash}p{0.7cm} 
        >{\centering\arraybackslash}p{0.7cm} 
        >{\centering\arraybackslash}p{1.5cm} 
        >{\centering\arraybackslash}p{1.3cm} 
        >{\centering\arraybackslash}p{1.3cm}}
        \toprule
        \multicolumn{2}{c}{\textbf{Categories}} & \multirow{2}{*}{\textbf{Datasets}} & \multirow{2}{*}{\textbf{Year}} & \multirow{2}{*}{\makecell{\textbf{ECT}}} & \multirow{2}{*}{\textbf{Emotion}} & \multirow{2}{*}{\makecell{\textbf{Training} \\ \textbf{Numbers}}} & \multirow{2}{*}{\makecell{\textbf{Testing} \\ \textbf{Numbers}}} \\ 
        \cline{1-2}
        \textbf{Modality} & \textbf{Scene} & & & & & & \\
        \midrule
        \multirow{16}{*}{\makecell{\vspace{2pt}Image\\-based \\ SFER \\Datasets}} & \multirow{5}{*}{\makecell{\vspace{2pt}Lab}} & \href{http://www.kasrl.org/jaffe.html}{JAFFE}~\cite{lyons1998coding} & 1998 & P & Sev & 213 & 213 \\
        & & \href{http://vasc.ri.cmu.edu/idb/html/face/facial_expression/}{CK+}~\cite{lucey2010extended} & 2010 & P/I & Sev & 241 & 241 \\
        & & \href{https://mmifacedb.eu/}{MMI}~\cite{valstar2010induced} & 2010 & P & Sev & 370 & 370 \\
        & & \href{http://www.cse.oulu.fi/CMV/Downloads/Oulu-CASIA}{Oulu-CASIA}~\cite{zhao2011facial} & 2011 & P & Sev & 720 & 240 \\
        & & \href{https://rafd.socsci.ru.nl/?p=main}{RaFD}~\cite{langner2010presentation} & 2010 & P & Sev, C & 1,448 &  160 \\
        \cline{2-8}
        & \multirow{6}{*}{\makecell{\vspace{2pt}Wild}} & \href{https://medium.com/@birdortyedi_23820/deep-learning-lab-episode-3-fer2013-c38f2e052280}{FER-2013}~\cite{goodfellow2013challenges} & 2013 & P/I & Sev & 28,709 & 3,589 \\
        & & \href{https://cs.anu.edu.au/few/}{SFEW 2.0}~\cite{dhall2011static} & 2011 & P/I & Sev & 958 & 436 \\
        & & \href{http://www.pitt.edu/~emotion/downloads.html}{EmotioNet}~\cite{fabian2016emotionet} & 2016 & P/I & Sev, C & 80,000 & 20,000 \\
        & & \href{http://www.whdeng.cn/raf/model1.html}{RAF-DB}~\cite{li2017reliable} & 2017 & P/I & Sev, Com & 12,271 & 3,068 \\
        & & \href{http://mohammadmahoor.com/affectnet/}{AffectNet}~\cite{mollahosseini2017affectnet} & 2017 & P/I & Sev, Con. & 283,901 & 3,500 \\
        & & \href{https://mmlab.ie.cuhk.edu.hk/projects/socialrelation/index.html}{ExpW}~\cite{zhang2018facial} & 2017 & P/I & Sev & 75,048 & 16,745 \\
        \cline{2-8}
        & \multirow{3}{*}{\makecell{\vspace{2pt}Lab \\ (3D)}} & \href{https://www.cs.binghamton.edu/~lijun/Research/3DFE/3DFE_Analysis.html}{BU-3DFE}~\cite{cheng20184dfab} & 2006 & P & Sev & 2,000 & 500 \\
        & & \href{http://bosphorus.ee.boun.edu.tr/}{Bosphorus}~\cite{savran2008bosphorus}  & 2008 & P & Sev & 2,326 & 2,326 \\
        & & \href{https://ibug.doc.ic.ac.uk/resources/4dfab/}{4DFAB}~\cite{cheng20184dfab} & 2018 & P/I & Sev & 1,440k & 360k \\
        \midrule
        \multirow{8}{*}{\makecell{\vspace{2pt}Video-\\based \\ DFER \\Datasets}} & \multirow{3}{*}{\makecell{\vspace{2pt}Lab}} & \href{http://vasc.ri.cmu.edu/idb/html/face/facial_expression/}{CK+}~\cite{lucey2010extended} & 2010 & P/I & Sev & 241 & 241 \\
        & & \href{https://mmifacedb.eu/}{MMI}~\cite{valstar2010induced} & 2010 & P/I & Sev & 1,450 & 1,450 \\
        & & \href{http://www.cse.oulu.fi/CMV/Downloads/Oulu-CASIA}{Oulu-CASIA}~\cite{zhao2011facial} & 2011 & P & Six & 2,160 & 720 \\
        \cline{2-8}
        & \multirow{5}{*}{\makecell{\vspace{2pt}Wild}} & \href{https://cs.anu.edu.au/few/AFEW.html}{AFEW 8.0}~\cite{dhall2018emotiw} & 2011 & P/I & Sev & 773 & 383 \\
        & & \href{https://caer-dataset.github.io/}{CAER}~\cite{lee2019context} & 2019 & P/I & Sev & 9,240 & 2,640 \\
        & & \href{https://dfew-dataset.github.io/}{DFEW}~\cite{jiang2020dfew} & 2020 & P/I & Sev & 12,000 & 3,000 \\
        & &  \href{https://wangyanckxx.github.io/Proj_CVPR2022_FERV39k.html}{FERV39k}~\cite{wang2022ferv39k} & 2022 & P/I & SE & 35,887 & 3,000 \\
        & & \href{https://mafw-database.github.io/MAFW/}{MAFW}~\cite{liu2022mafw} & 2022 & P/I & Sev, C, A, D, H, Com & 8,036 & 2,009 \\
        \bottomrule
    \end{tabular}
}
\vspace{-0.5cm}
\label{tab:summary_datasets}
\end{table}

\subsection{Video-based DFER Datasets}
The video-based DFER datasets are typically composed of videos or image sequences with durations ranging from 0.4 to 5 seconds, which are also divided into two categories: controlled laboratory scenes and complex real scenes.

For in-the-lab DFER, there are 3 widely used as benchmark datasets, including 1) \textbf{CK+}~\cite{lucey2010extended} comprises 593 facial expression sequences posed by 123 subjects, of which only 327 sequences are labeled with seven basic emotions and contempt; 2) \textbf{MMI}~\cite{valstar2010induced} consists of 740 images and 2,900 video sequences, which were posed by 32 subjects in a laboratory setting, depicting seven basic emotions; 3) {Oulu-CASIA}~\cite{zhao2011facial} includes 2,880 video sequences captured in a laboratory setting with 80 subjects posing in front of the camera with the six basic facial expressions.

For in-the-wild DFER, there are 5 widely used as benchmark datasets, including 1) \textbf{AFEW 8.0}~\cite{dhall2018emotiw}, used in the EmotiW competition, is a multimodal video dataset featuring spontaneous human expressions collected from TV and film clips, encompassing various head poses, object occlusions, and lighting conditions; 2) \textbf{CAER}~\cite{lee2019context} contains 13,201 annotated facial video clips from American television series; 3) \textbf{DFEW}~\cite{jiang2020dfew} offers 12,059 clips from 16,372 clips annotated via a voting mechanism, segmented into five sections for cross-validation; 4) \textbf{FERV39k}~\cite{wang2022ferv39k} contains 38,935 video segments from 22 scenes and four situations, annotated with seven basic expressions, covering diverse scenarios and expression intensities; 5) \textbf{MAFW}~\cite{liu2022mafw} is a large-scale, multi-modal affective dataset with 10,045 video and audio clips from over 1,600 movies and TV dramas, categorized into eleven emotions, including seven basic and four additional expressions.

\subsection{Evaluation Metrics}

When evaluating the performance of FER models, several key evaluation metrics are commonly used: 1) \textbf{Accuracy} refers to the ratio of the number of correctly predicted samples to the total number of samples, which measures the FER model's ability to correctly identify expressions. 2) \textbf{Recall} measures the proportion of correctly predicted samples for a particular expression relative to the total actual samples of that expression; 3) \textbf{Weighted Average Recall (WAR)} calculates the weighted average by multiplying the recall of each class by its proportion in the dataset, providing a balanced view of performance across different classes; 4) \textbf{Unweighted Average Recall (UAR)} averages the recall rates across all classes without considering class proportions, offering a fair assessment of model performance in imbalanced datasets. Besides, the \textbf{Confusion Matrix} is a two-dimensional grid that visualizes a FER model's prediction results by comparing actual versus predicted categories.

\begin{figure*}[ht]
\vspace{-0.1cm}
  \centering
   \includegraphics[width=0.95\linewidth]{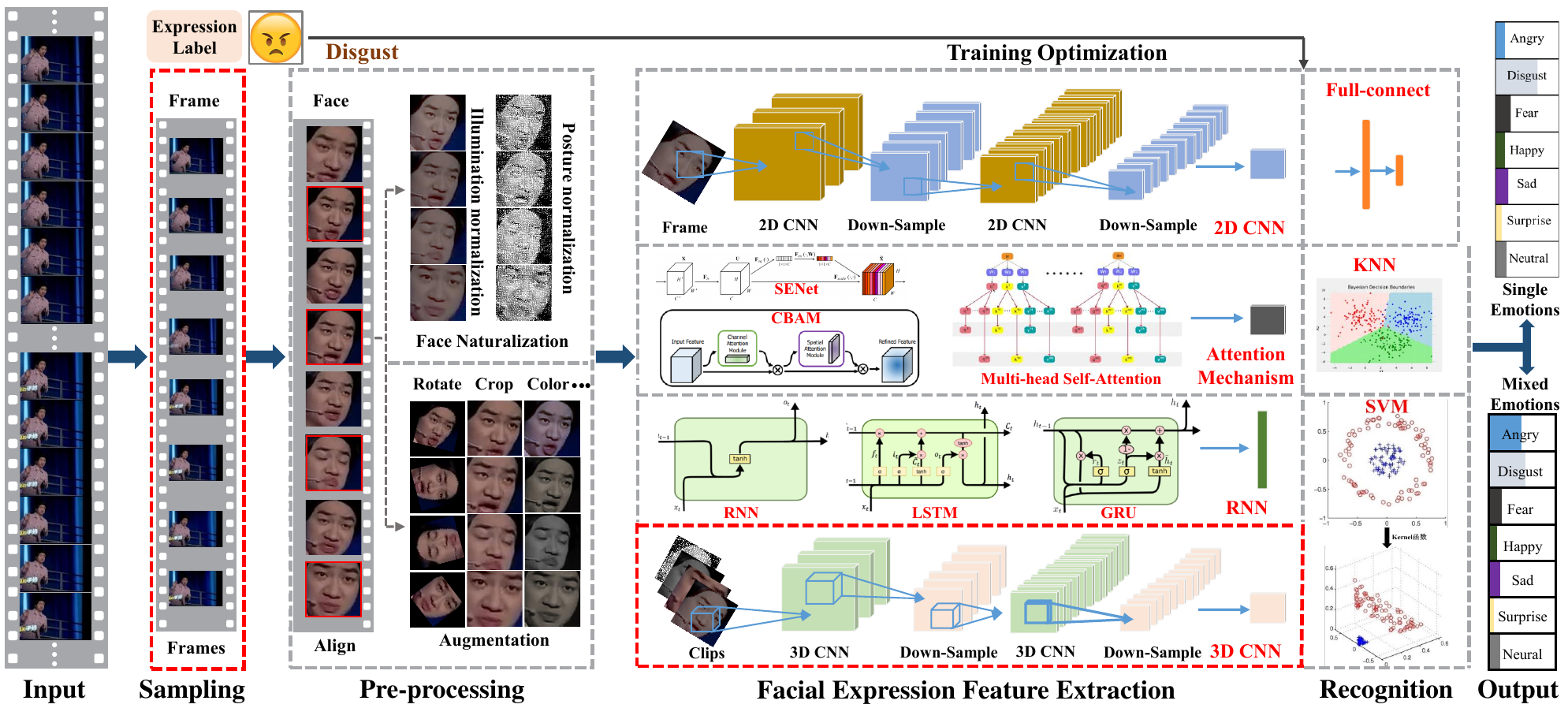}
   \vspace{-0.2cm}
   \caption{The workflow and main components of generic facial expression recognition.}
   \label{fig:fig_workflow}
   \vspace{-0.5cm}
\end{figure*}


\section{A Workflow of Generic FER}
\label{sec:Tutorial}

Fig.~\ref{fig:fig_workflow} shows the workflow and main components of generic FER. Four critical steps are typically included: 1) acquiring and sampling strategy of dynamic facial expression image sequences (only for DFER in red dashed rectangle); 2) pre-processing of align, naturalization and augmentation; 3) facial expression feature extraction of 2D CNN, Attention, RNN and 3D CNN (only for DFER in red dashed rectangle); and 4) expression recognition with single or mixed emotion.



\subsection{Facial Frame Sampling}

Since dynamic facial expressions in natural multi-scene environments usually last between 0.5 and 4 seconds~\cite{ben2021video}, researchers manually crop video clips at this interval when constructing DFER datasets. The number of frames transmission per second is usually 24 or higher. The step of "Sampling" in Fig.4 shows an example of 8 frames, which are sampled from a 40-frame disgust video sequence. There are two common sampling methods: uniform frame sampling and random frame sampling. In DFEW~\cite{jiang2020dfew} and FERV39k~\cite{wang2022ferv39k}, the uniform sampling strategy is first used to generate a sequence face images of length 16 or 8 from all available video clips with the assistance of random sampling and Time Interpolation Method (TIM). In \cite{zhao2021former,li2023intensity}, the facial expression video is evenly segmented into $U$ segments, then $V$ image frames are randomly selected from each segment, eventually, the length of a facial image sequence is $U \times V$. Note facial frame sampling is only used for dynamic emotion data (video-based DFER tasks).

\subsection{Facial Data Preprocessing}

The original dynamic facial expression sequences obtained from the natural world often contain expression-irrelevant variables, such as complex backgrounds, varying illumination, and face pose changes. To exclude the irrelevant information, serialized face image preprocessing~\cite{ranjan2017hyperface} is necessary to align, normalize, and augment the semantic information of facial regions before deep feature extraction.

\textbf{Face Alignment} focused on the automatic detection of facial landmarks to eliminate background and non-expression elements, which can be generally categorized into two main approaches: cascaded regression (CPR) models and DL-based methods. On the basis of the CPR, Robust CPR (RCPR)~\cite{xiong2013supervised} improved robustness against occlusions and shape variations.  Early methods, such as DCNN~\cite{sun2013deep}, and TCNN~\cite{wu2017facial}, directly applied multi-layer CNNs to learn key features of  face (facial landmarks) for alignment. Recently, SfSNet \cite{Kar_2023_BMVC} leveraged DL-based models for enhanced performance. In addition, the landmark strategy is improved from the perspective of semantic understanding. Zhou et al. \cite{jirr_10205248} proposed the Self-adaptive Ambiguity Reduction (STAR) loss to address semantic ambiguity in landmark detection.

\textbf{Face Normalization} involves illumination normalization ~\cite{hu2021fin} and pose normalization \cite{tran2017disentangled}. Ma et al.~\cite{ma2018face} utilized cyclic consistency loss for light normalization as a style transfer problem, while Han et al. \cite{han2019asymmetric} developed Asymmetric Joint GANs  for controlled reillumination. Disentangled Representation learning-GAN \cite{tran2018representation} focused on achieving pose-invariant facial representations via GAN-based models. Tripathy et al. \cite{tripathy2020icface} introduced a self-supervised approach for dynamic face reproduction. Unlike methods that rely on separate classifiers for each pose, Zhang et al. \cite{zhang2020geometry} proposed an end-to-end deep GAN model that integrates face synthesis and pose-invariant expression recognition.

\textbf{Data Augmentation} can effectively provide sufficient samples when the limited labeled facial expression images or sequences in the dataset cannot meet the requirement of FER training. It can be divided into two types: offline and on-the-fly. Scaling and tilting the original images, along with the inclusion of random noise and scrambling, are the most commonly employed offline data augmentation techniques~\cite{lopes2017facial}. GAN-based data synthesis methods \cite{zhang2020geometry, sahu2020modeling} could also be applied to generate various face and expression images. The on-the-fly data augmentation \cite{li2021adaptively} usually use data expansion methods embedded in DL toolkits, such as random rotate and crop, and color jitter. 


\subsection{Facial Emotion Feature Extraction} 
As deep learning continues to advance for various tasks, especially image and video related tasks, recent efforts in FER have concentrated on optimizing network architectures applied for facial emotion feature extraction. These network models can be classified into four categories: 1) deep convolutional neural networks; 2) attentional mechanisms; 3) recurrent neural networks, and 4) 3D convolutional neural networks. Specific network structures will be described in Sec.~\ref{sec:Static} and Sec.~\ref{sec:Dynamic} according to the task challenges.

\textbf{CNN-based Models} have achieved tremendous success in the field of computer vision, particularly excelling in tasks such as image classification, object detection, and segmentation. By utilizing convolutional layers and pooling layers, CNNs such as VGGNet \cite{simonyan2015a} and ResNet \cite{He_2016_CVPR} can effectively extract both local and global features from images, making them suitable for static facial expression feature extraction.

\textbf{Attention-based Models} help the network model extract more representative information by adaptively finding spatial regions, channels, or Spatio-temporal sequences that are meaningful to the task based on the input feature vectors. The Vision Transformer (ViT) \cite{dosovitskiy_image_2021} excels in capturing long-range dependencies in sequence data and images using self-attention or multi-Head attention mechanisms, making them suitable for complex FER tasks~\cite{li2022affective, lishutao_10095448_2023}.

\textbf{Recurrent Neural Networks (RNNs)} learn mappings between complex feature tensors and combine temporal and spatial information (with CNNs) to further improve performance. Mostly used RNN-architecture networks are Long Short-Term Memory Networks (LSTM) \cite{RNN_1997} and Gated Recurrent Unit Networks (GRU) \cite{cho-etal-2014-learning}. They can effectively capture subtle changes in facial expressions regarding image patches or video frames as a time series~\cite{liu2020saanet}.

\textbf{3D Convolutional Neural Networks} (C3D) \cite{ji20123d} model both spatial and temporal signals simultaneously . 3D CNNs \cite{tran2015learning, 10023947_dr} extend traditional 2D convolutional operations  like ResNet or Inception \cite{szegedy2016rethinking, he2016deep} into the temporal dimension by applying 3D convolutions. These C3Ds can model the spatiotemporal patterns of facial movements, making them particularly effective in capturing the nuances of dynamic expressions~\cite{jiang2020dfew, wang2022ferv39k}.

\subsection{Recognition of Facial Emotions}

Recognition of facial emotions aims to calculate the classification probability of input data by traditional machine learning or deep learning methods, and eventually determine the expression with single labels or mixed labels.

\textbf{Machine Learning based Classifier} includes the widely accepted traditional machine learning classification methods, such as Support Vector Machine (SVM) \cite{azazi2015towards} and Multi-layer Perceptron (MLP) \cite{akhtar2020intense}. Among them, SVM is one of the most effective classifiers for affective computing.

\textbf{Deep Learning based Classifier} is often used in DL-based frameworks \cite{jin2021learning, kaneko2017generative}, which classifies expressions with the features extracted by previous network layers. Specifically, DL-based networks, such as ResNet \cite{He_2016_CVPR} and 3D Convolutional Neural Network (C3D) \cite{tran2015learning}  consider facial emotion features for end-to-end expression recognition of images or sequences by fully-connected layers.

\section{Image-based Static FER}
\label{sec:Static}

Image-based static facial expression recognition (SFER) involves extracting features from a single image, which captures complex spatial information that related to facial expressions, such as landmarks, and their geometric structures and relationships. In the following, we will first introduce the general architecture of SFER, and then elaborate specific design of SFER methods from the challenge-solving perspectives, including disturbance-invariant SFER, 3D SFER, uncertainty-aware SFER, compound SFER, cross-domain SFER, waek-supervised SFER, and cross-modal SFER.






\subsection{General SFER}
\label{sec:Static_general}

\begin{figure}[t]
\vspace{-0.1cm}
  \centering
   \includegraphics[width=1.0\linewidth]{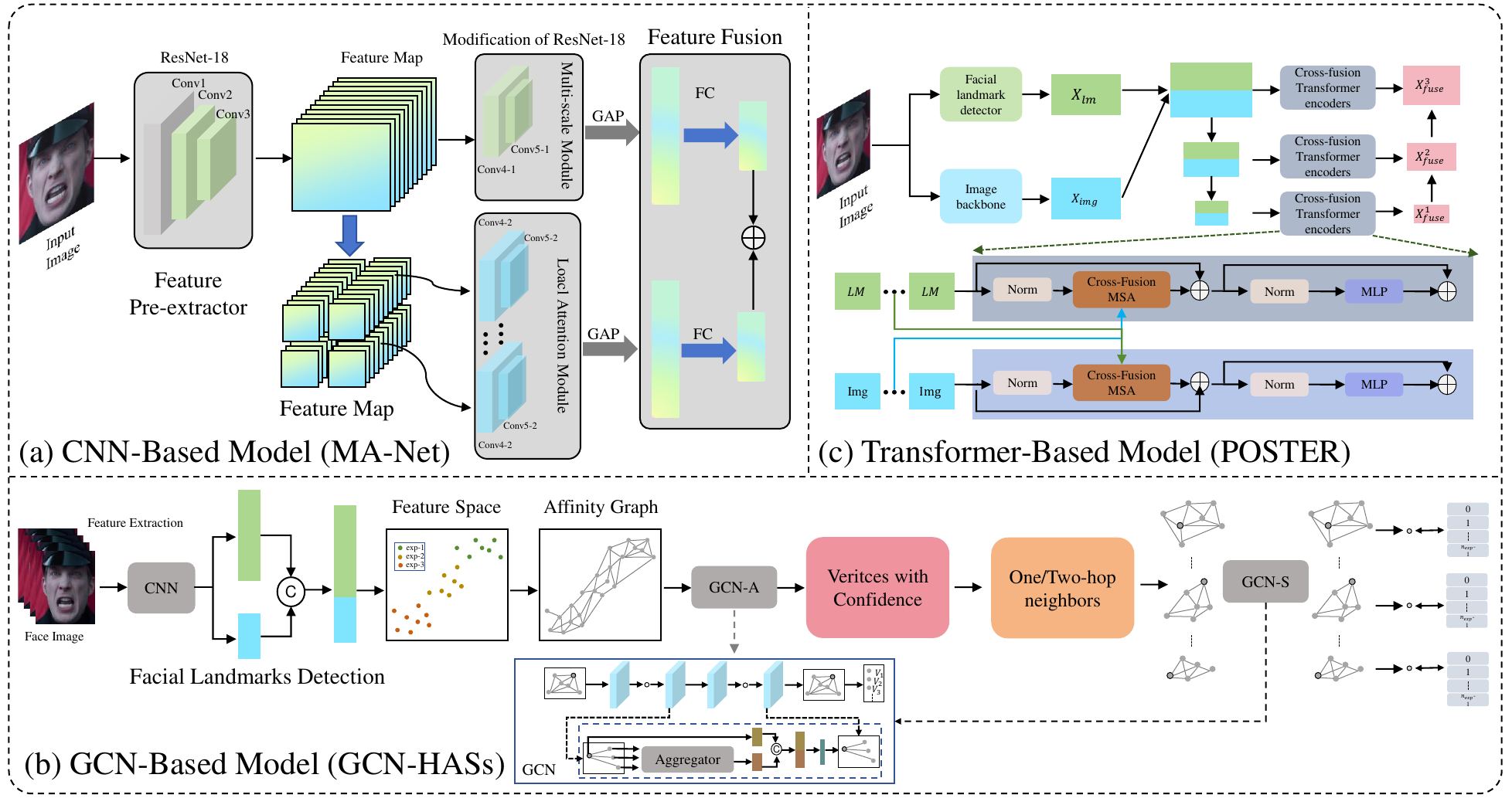}
   \vspace{-0.5cm}
   \caption{The architecture of general SFER. Figure is reproduced based on (a) CNN-based model \cite{zhao2021learning}, (b) GCN-based model~\cite{Liu10173748}, and (c) Transformer-based model~\cite{Chen_10350905}.}
   \label{fig:fig_general}
   \vspace{-0.3cm}
\end{figure}

A general SFER  often involves global and local or multi-scale feature extractions, feature fusion, and emotion classification. Fig.~\ref{fig:fig_general} shows an example architecture of general SFER. In this process, deep learning models serve as the foundational framework, mainly including Convolutional Neural Networks (CNNs), Graph Convolutional Networks (GCNs), and Transformer-based models. The integration and advancement of these deep learning architectures have significantly enhanced the performance of SFER systems, enabling robust recognition across diverse environments.


\subsubsection{CNN-based Models}
CNN-based methods~\cite{cai2018island, 9244224, yan2020low} have proven instrumental in SFER by efficiently extracting  local and global facial characteristics through layered convolution and pooling operations, facilitating accurate expression classification. Facial Motion Prior Networks (FMPN) \cite{8965826} and Oriented Attention Pseudo-Siamese Network (OAENet) \cite{wang2021oaenet} leveraged convolutional blocks to capture global and local facial information via facial landmarks and correlation coefficients. As shown in Fig.~\ref{fig:fig_general}(a), the global multi-scale and local attention network (MA-Net) \cite{zhao2021learning} utilized multi-scale module and a local attention module to extract both local and global facial features.


\subsubsection{GCN-based Models}
Compared with traditional CNN approaches, Graph Convolutional Network (GCN)-based methods are particularly better at handling the geometric relationships and topological information of facial features to capture spatial dependencies through graph structures for recognizing subtle changes in expressions. As shown in Fig.~\ref{fig:fig_general}(b), Liu et al.~\cite{Liu10173748} developed a method that utilizes high aggregation subgraphs (GCN-HASs) for FER. By emphasizing the importance of high-order neighbors and employing vertex confidence, their approach constructs subgraphs that effectively capture the intricate relationships between facial expressions, leading to significant improvements in both recognition accuracy and efficiency. These contributions highlight the innovative advancements in leveraging GCNs for SFER, demonstrating their potential to surpass the limitations of CNN-based methods. Jin et al.~\cite{jin2021learning} employed a graph-structured representation (DDRGCN) where each node corresponds to appearance information around facial landmarks, and edges encode the geometric relationships between these nodes. This approach captures both local appearance and spatial geometry, providing a robust framework for recognizing facial expressions.


\subsubsection{Transformer-based Models}
 Since Transformer-based methods can capture global information and spatio-temporal relationships in facial expressions~\cite{xue2021transfer, li2022affective}, novel architectures are optimized by introducing multi-scale and cross-modal attention mechanisms. As shown in Fig.~\ref{fig:fig_general}(c), a Pyramid Cross-Fusion Transformer network (POSTER) \cite{Chen_10350905} utilized a transformer-based cross-fusion approach to effectively integrate facial landmark features with image features, directing attention to important facial regions and enhancing scale invariance. In addition, Li et al.\cite{li2022affective} employed a Masked Auto-Encoder pretrained on unlabeled face images, combined with a pretrained Vision Transformer and CNN, to tackle the issue of limited annotations in facial expression data for affective behavior analysis.

\begin{figure}[t]
\vspace{-0.1cm}
  \centering
   \includegraphics[width=1.0\linewidth]{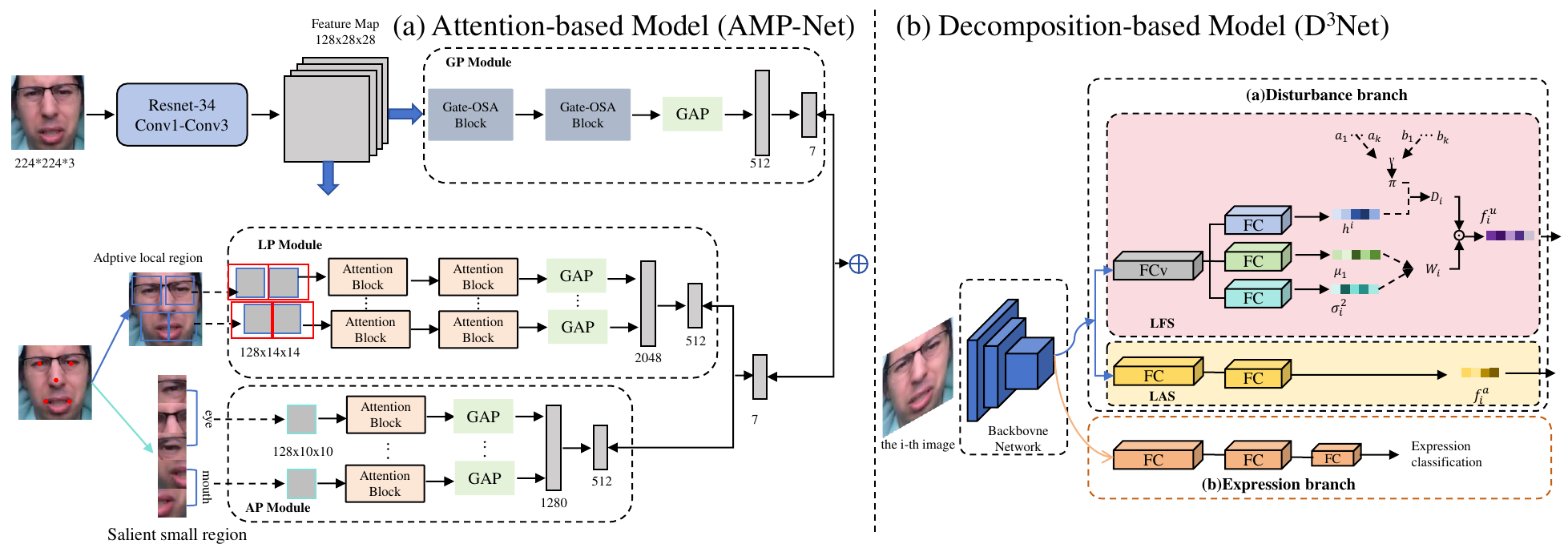}
   \vspace{-0.5cm}
   \caption{The architecture of disturbance-invariant SFER. Figure is reproduced based on (a) Attention-based model~\cite{Liu2022Ada} and (b) Decomposition-based model~\cite{mo2021d3net}.}
   \label{fig:Figue_Disturbance}
   \vspace{-0.3cm}
\end{figure}

\subsection{Disturbance-invariant SFER}
\label{sec:Static_Disturbance}

One of the main challenges in FER is to address the disturbance caused by various disturbing factors~\cite{Lee_2023_ICCV, liu2023patch, wang2024mgr, mo2021d3net}, including common ones (such as identity, pose, and illumination) and potential ones (such as hairstyle, accessory, and occlusion). These disturbing factors will lead to partial information missing. To overcome the impact of disturbance, it is critical to extract effective facial expression features from available facial regions. 


\subsubsection{Attention-based Models}
 The attention-based models~\cite{9511448_FER, liu2023patch} based on attention mechanism ~\cite{Wangfy9496600} can help the model better focus on the unoccluded facial regions, thereby improving the accuracy and robustness of expression recognition in complex backgrounds and lighting conditions. 

 \textbf{Region-based FER methods} analyze a face image by dividing it into overlapping or non-overlapping local regions, allowing the model to concentrate on localized features for more precise expression recognition. Li et al.~\cite{19_li2018occlusion} introduced the Patch-Gated Unit (PG-Unit), which computes one-dimensional weights for regions of interest based on facial landmarks. These weights are then applied across feature dimensions using self-attention and relation-attention modules, enhancing the model's ability to focus on the most relevant facial regions. 

 \textbf{Holistic-region-based methods} often have two branches to extract global and local features. For example, Wang et al. \cite{30_wang2021local} proposed the local attention module and correlation attention learning to obtain local attention maps and an overall saliency feature. Similarly, Fig.~\ref{fig:Figue_Disturbance}(a) presented an adaptive multilayer perceptual attention network~\cite{Liu2022Ada}, which extracted global, local, and salient facial emotional features by incorporating various fine-grained features. These approaches aims to learn the underlying diversity and crucial information inherent in facial expressions.

 \begin{figure}[t]
\vspace{-0.1cm}
  \centering
   \includegraphics[width=1.0\linewidth]{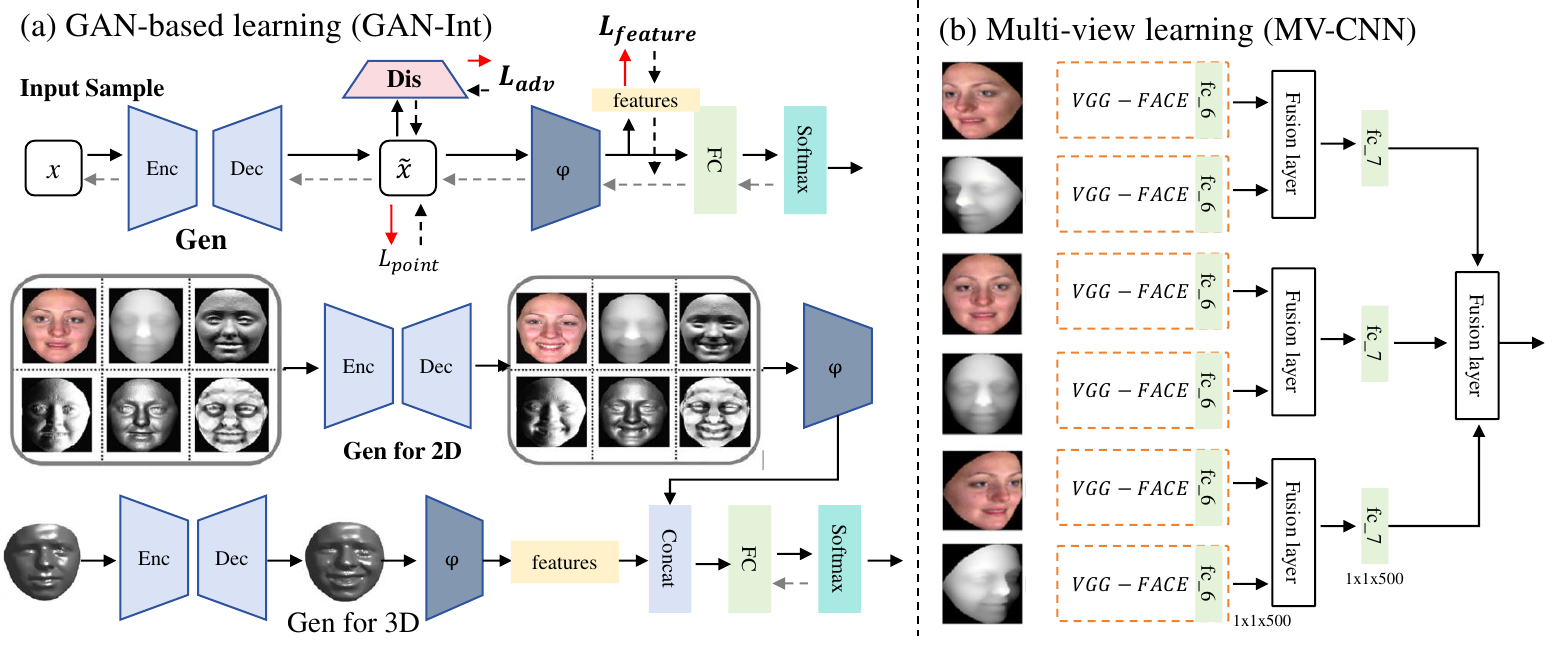}
   \vspace{-0.5cm}
   \caption{The architecture of 3D SFER. Figure is reproduced based on (a) GAN-based learning (GAN-Int)~\cite{yang2021intensity} and (b) Multi-view learning (MV-CNN)~\cite{vo20193d}.}
   \label{fig:Figue_3D}
   \vspace{-0.3cm}
\end{figure}



\subsubsection{Decomposition-based Models}

Decomposition-based models~\cite{jiang2022disentangling, Yanyan_fer, mo2021d3net} aim to disentangle facial expressions from identity and posture, generating discriminative facial expression features. As shown in Fig.~\ref{fig:Figue_Disturbance}(b), the dual-branch disturbance disentangling network (D\textsuperscript{3}Net)~\cite{mo2021d3net} includes both an expression branch and a disturbance branch. The disturbance branch is divided into a label-aware sub-branch (LAS) that captures common disturbances through transfer learning, and a label-free sub-branch (LFS) that encodes potential disturbances using an unsupervised Indian Buffet Process (IBP) prior. Adversarial training is employed to further separate disturbance features from expression features, improving feature disentanglement. The feature decomposition and reconstruction learning (FDRL)~\cite{ruan2021feature} integrates a feature decomposition network to model similarities and a feature reconstruction network to capture relationships and reconstruct expression features using intra- and inter-feature relation modules.  Additionally, Latent-OFER~\cite{Lee_2023_ICCV} detected occlusions and reconstructing missing regions using latent vectors from unoccluded patches by the effect of decomposition-based models in isolating and amplifying expression-specific features.

\subsection{3D SFER}
\label{sec:Static_3D}

Despite significant advances achieved in 2D FER, it is still difficult to distinguish some facial muscle action units in 2D images due to limitations such as lighting conditions, poses, and makeup. Since 3D facial shape models include depth information and enable the observation of facial feature changes from multiple angles, 3D FER works~\cite{zhu2023var, liu2023joint} utilized complementary and redundant information in 2D and 3D data to capture subtle deformations and details.


\subsubsection{GAN-based Learning}
Generative Adversarial Network (GAN)-based methods can generate high-quality, diverse, and nearly-realistic facial expression images through the adversarial training of generators and discriminators. This make it easier for improving the generalization ability of FER models. As shown in Fig.~\ref{fig:Figue_3D}(a), Yang et al. utilized GAN model (GAN-Int)~\cite{yang2021intensity} to jointly design intensity enhancement and expression recognition, ensuring that synthesized faces exhibit high-intensity expressions. Similarly, Zhang et al. \cite{zhang2018joint} proposed Joint Pose and Expression GAN-based model (JPE-GAN) to simultaneously perform facial image generation and pose-invariant FER by corporately utilizing different poses and expressions.

\subsubsection{Multi-view Learning}

Multi-view learning in 3D FER \cite{yanbo2023cc,liu2023joint} utilized multi-angle 3D facial images and combines features from various perspectives to distinguish different expressions, effectively addressing variations in pose and lighting conditions, thus enhancing overall recognition performance. As shown in Fig.~\ref{fig:Figue_3D}(b), Vo et al. \cite{vo20193d} proposed a novel multi-view CNN model (MV-CNN), which incorporates multi-view facial images and facial prior information for 3D FER. In addition, the joint spatial and scale attention network (SSA-Net)~\cite{liu2023joint} localized proper regions for simultaneous head pose estimation and FER. The SSA-Net uses spatial attention to identify expression-relevant regions at various scales and employs scale attention to select the most informative scales, learning pose-invariant and expression-discriminative representations.

\subsection{Uncertainty-aware SFER}
\label{sec:Static_Uncertainty}

FER tasks are inherently challenged by factors such as image quality, facial posture, and lighting conditions, which further introduce data and label uncertainty~\cite{she2021dive}. Uncertainty-aware SFER models aim to classify the facial expressions while handling the uncertainty of each class.


\begin{figure}[t]
\vspace{-0.1cm}
  \centering
   \includegraphics[width=1.0\linewidth]{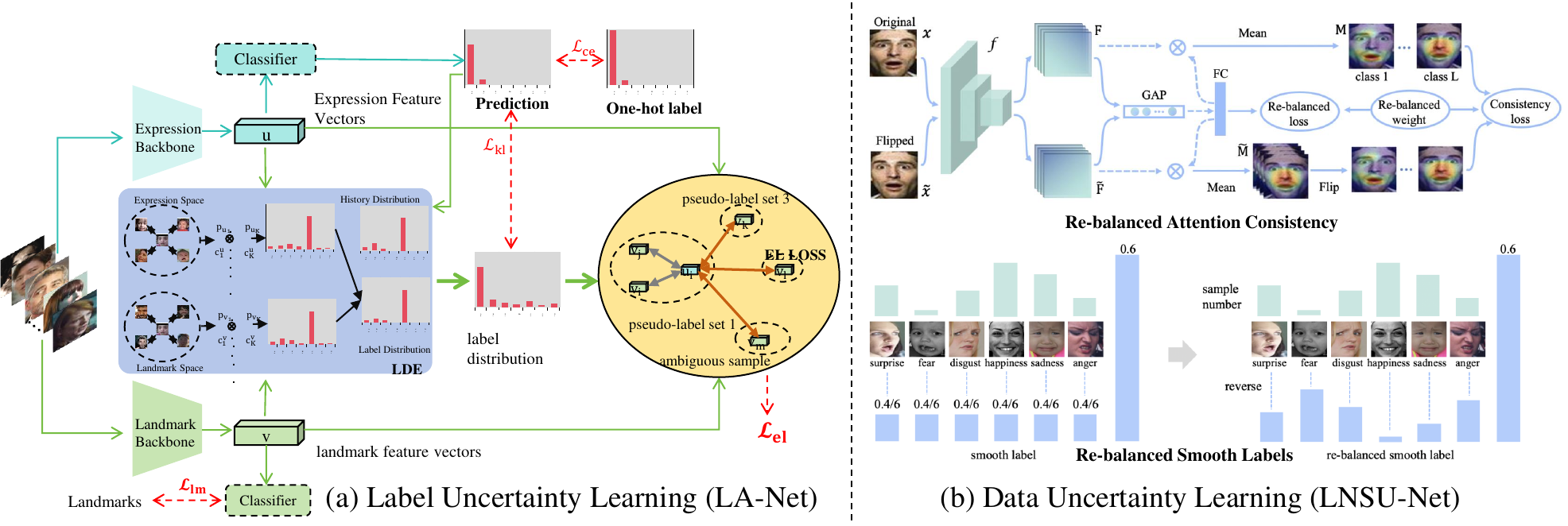}
   \vspace{-0.5cm}
   \caption{The architecture of uncertainty-aware SFER. Figure is reproduced based on (a) the label uncertainty learning (LA-Net)~\cite{wu2023net} and (b) data uncertainty learning (LNSU-Net)~\cite{zhang2024leave}.}
   \label{fig:uncertainty}
   \vspace{-0.3cm}
\end{figure}

\subsubsection{Label Uncertainty Learning}
Label data may contain noise or errors~\cite{chen2020label,Deng2022learn} due to human annotation mistakes or inherent data ambiguity, significantly affecting model performance. Robust techniques have been used to deal with noise labels: 1) design Label Distribution Learning on Auxiliary Label Space Graphs (LDL-ALSG)~\cite{chen2020label} to suppress noise; 2) use unlabeled data to assist the model in recognizing and correcting noise in label data \cite{Xucs_9403940}; 3) reduce the impact of noisy labels on FER models by erase attention consistency (ECA)~\cite{Deng2022learn}, similarly, as shown in Fig.~\ref{fig:uncertainty}(a), LA-Net \cite{wu2023net} also leveraged facial landmarks for attention and label correction to counter label noise.

\subsubsection{Data Uncertainty Learning}

Large-scale FER datasets collected in the wild often encounter issues like image blur, noise, and low resolution, leading to ambiguity in emotion recognition~\cite{wang2020suppressing}. These challenges complicate distinguishing between images with multiple emotions and those with noisy labels. To address this, Zhang et al.\cite{zhang2021relative} introduced a relative uncertainty learning framework that estimates the uncertainty of each prediction relative to others, improving model robustness. The Emotion Ambiguity-Sensitive Cooperative Networks (EASE)\cite{Yang2022ease} further tackle this by categorizing training samples into clean, noisy, and conflict groups, enhancing network diversity and representation learning. Incorporating auxiliary tasks, Zhao et al.\cite{Zhaogy_2023uncertain} developed an uncertainty-aware model using multi-task auxiliary correction to improve FER accuracy under uncertain conditions. As shown in Figure~\ref{fig:uncertainty}(b), Deng et al.\cite{zhang2024leave} employed re-balanced attention maps, enabling models to better extract information from underrepresented classes like fear or disgust, thus enhancing overall performance in FER.

\subsection{Compound SFER}
\label{sec:Static_Compound}

Compound emotions~\cite{li2017reliable} refer to complex emotional states formed by the combination of at least two basic emotions, which are not independent, discrete categories. Compared with seperate basic emotions, compound emotions are more capable of representing the diversity and continuity of human's complex emotions. Li et al. \cite{9511468_multilabel} proposed a self-supervised exclusive-inclusive interactive learning method for multi-label FER, effectively capturing and disentangling both inclusive and exclusive facial expressions within a single image. Deng et al.~\cite{jiang2023improving} improved multi-label FER by introducing attention flipping consistency loss and label-guided spatial attention dispersing loss, which bolstered network stability, interpretability, and performance without additional data. Additionally, Deng et al.~\cite{jiang2024joint} addressed basic-compound FER as a single-label multi-class task, proposing the iterated soft label mining algorithm and expression correlation score learning loss to effectively leverage label correlations.

\begin{figure}[t]
\vspace{-0.1cm}
  \centering
   \includegraphics[width=0.9\linewidth]{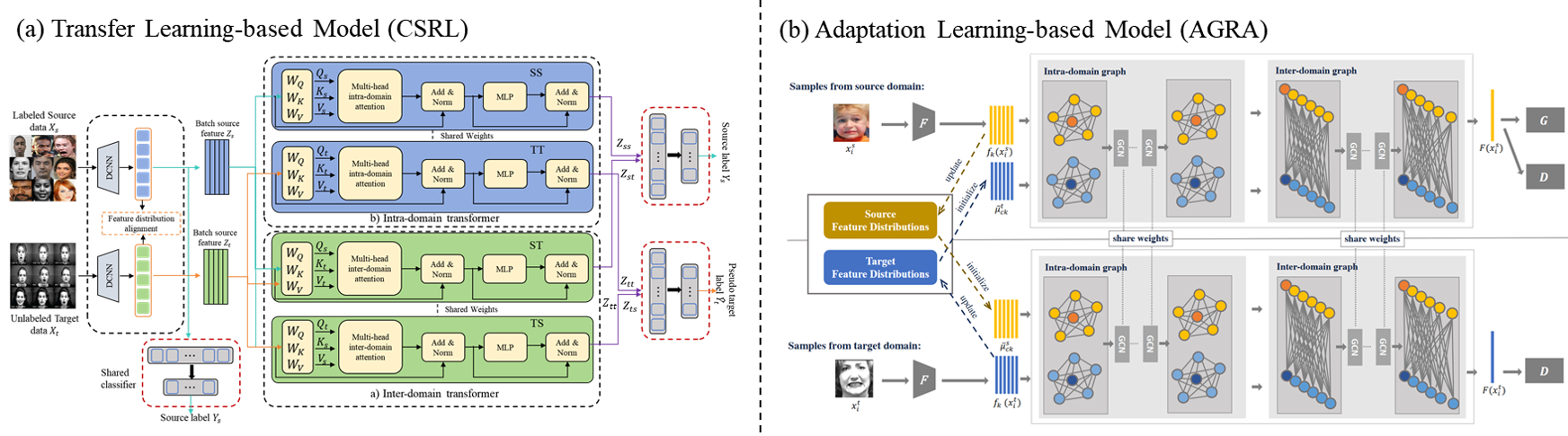}
   \vspace{-0.2cm}
   \caption{The architecture of cross-domain SFER. Figure is reproduced based on (a) the transfer learning-based model (CSRL)~\cite{chen2023cross} and (b) the adaption learning-based model (AGRA)~\cite{Chen_pami} .}
   \label{fig:fig_cross_domain_sfer}
   \vspace{-0.3cm}
\end{figure}

\subsection{Cross-domain SFER}
\label{sec:Static_Cross}

In real-world environments, facial expressions vary across race, culture, and age, as well as annotators' cultural and experiential biases, reducing the performance of existing recognition methods on diverse datasets~\cite{Chen_pami}. Fortunately, advancements~\cite{Wang_uda} in transfer learning and adaptation learning have facilitated the transfer of knowledge from labeled source domains to target domains, enhancing the generalization of the cross-domain SFER model across different contexts.

\subsubsection{Transfer Learning-based models}

Variations in data collection conditions across different datasets can lead to significant performance degradation when models trained on one dataset are applied to another. As shown in Fig.~\ref{fig:fig_cross_domain_sfer}(a), the Cross-domain Sample Relationship Learning (CSRL)~\cite{chen2023cross} reduces domain discrepancy by leveraging intrinsic sample relationships across domains. Specifically, during training, inter-domain sample transformers are designed to explore similarity relationships between source and target domains, while intra-domain sample transformers capture internal structures within each domain. Furthermore, a joint alignment strategy is employed to align feature distributions and sample relationships across domains, enhancing the model's generalization ability by aligning both local sample similarities and global domain distributions. Zheng et al. \cite{Zheng_9761954} proposed a joint local-global discriminative subspace transfer learning method that learns a domain-invariant subspace by integrating both local and global information. Additionally, Zheng et al. \cite{Zheng_7465718} introduced a cross-domain color FER method using transductive transfer subspace learning to identify a shared subspace for effective knowledge transfer. Further, Zheng et al. \cite{Zheng_10137555} proposed learning a common latent embedding space to enhance cross-domain FER. They also suggested learning transferable sparse representations for effective cross-corpus recognition \cite{Zheng_9423630}, aiming to extract discriminative features that can generalize across datasets.

\subsubsection{Adaptation Learning-based models}

Adaptation learning plays a pivotal role in addressing the domain shift challenges (different feature distribution of the same expression in different datasets) inherent in cross-domain FER~\cite{Chen_pami, 9676449}. Adversarial learning helps the model achieve domain adaptation between the source domain and the target domain via approximating the feature distributions of the source and target domains \cite{Zhao_9238468}. As shown in Fig.~\ref{fig:fig_cross_domain_sfer}(b), Chen et al. \cite{Chen_pami} combined graph representation propagation with adversarial learning for global-local feature co-adaptation across domains. Similarly, a Multi-source Adversarial Domain Aggregation Network (MADAN)~\cite{zhao_madan_2021} learnt domain-invariant features from multiple source domains for effective transfer to the target domain. To achieve cross-domain and discriminative feature representations, Li et al. \cite{8993748} introduced the deep Emotional Conditional Adaptation Network (ECAN), which aligns both marginal and conditional distributions across domains. Additionally, Gao et al. \cite{gao2023ssa} proposed multi-domain adaptive attention (SSA-ICL) with Intra-dataset Continual Learning, effectively adapting to multiple target domains and mitigating catastrophic forgetting. Recently, Zheng et al. \cite{Zheng2024graph} introduced a graph-diffusion-based domain-invariant representation learning, capturing the underlying manifold structure of facial expressions to achieve domain-invariant representations.

\subsection{Weak-supervised SFER}
\label{sec:Static_Weak}

\begin{figure}[t]
\vspace{-0.1cm}
  \centering
   \includegraphics[width=0.85\linewidth]{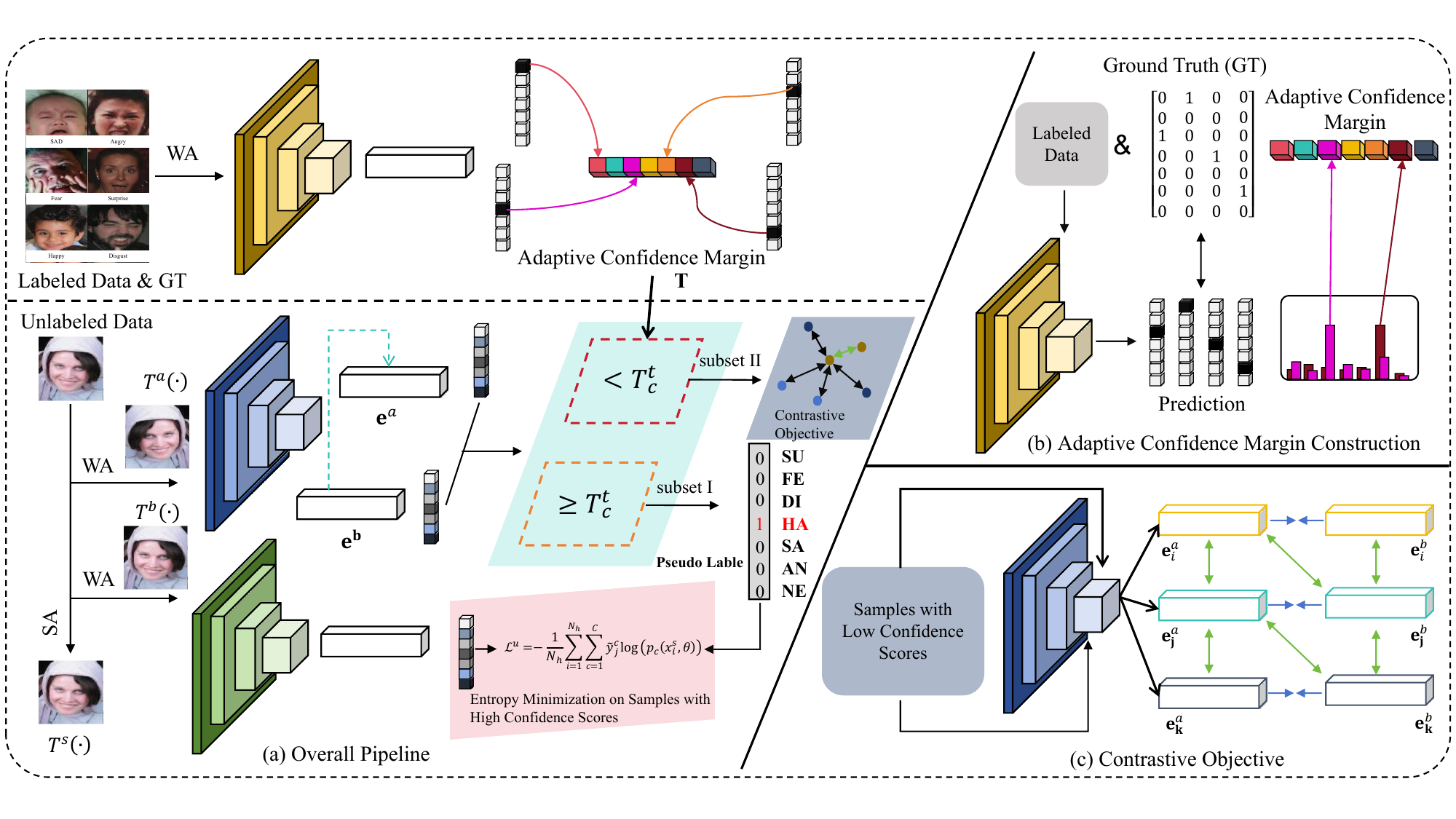}
   \vspace{-0.2cm}
   \caption{The architecture of weak-supervised SFER. Figure is reproduced based on the Ada-CM~\cite{li2022towards}.}
   \label{fig:semisupervisedsfer}
   \vspace{-0.5cm}
\end{figure}

Weak-supervised learning in SFER involves training models with scarce or partially available labeled data, leveraging both labeled and unlabeled data, to learn facial discriminative expression representation. In recent work, Zhang et al.~\cite{39_Zhang2020WeaklySL} advanced this field by weakly supervising local regions of interest and incorporating relational reasoning between local and global features. As shown in Fig.~\ref{fig:semisupervisedsfer}, the Adaptive Confidence Margin (Ada-CM)~\cite{li2022towards} leveraged and partitioned all unlabeled data into two subsets based on confidence scores: high-confidence samples used for pseudo-label matching, while low-confidence samples contributing to feature-level contrastive learning. Shu et al. further ~\cite{Shu_2022_BMVC} revisited contrastive learning in a semi-supervised context, proposing a framework that effectively utilizes unlabeled data to boost FER model performance. Recently, Liu et al. introduced weakly supervised contrastive learning (WSCFER)~\cite{liuhh_wscl} by integrating instance-level and class-level representation learning, which balances feature discrimination through contrastive learning and partial consistency loss, minimizing focus on irrelevant details.

\subsection{Cross-modal SFER}
\label{sec:Static_Crossmodal}

\begin{figure}[t]
\vspace{-0.1cm}
  \centering
   \includegraphics[width=0.85\linewidth]{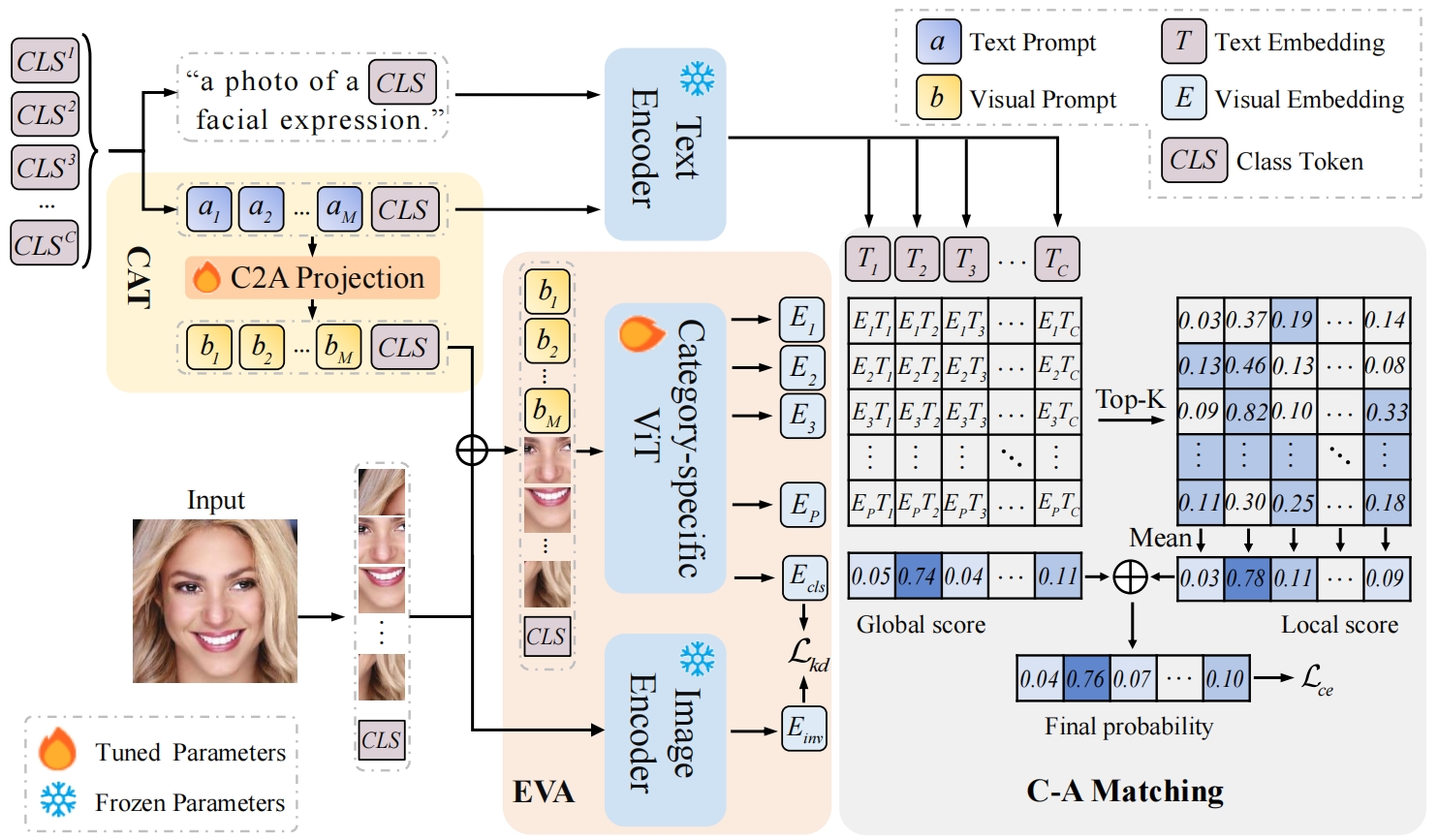}
   \vspace{-0.2cm}
   \caption{The architecture of cross-modal SFER. Figure is reproduced based on the CEprompt \cite{CEPrompt_zff_2024}.}
   \label{fig:figCEprompt}
   \vspace{-0.3cm}
\end{figure}

Cross-modal SFER methods~\cite{Zhang_2023_ICCV,lv2024visual} integrated the visual facial information with emotion conception  from textual sources using the visual language pre-training (VLP)~\cite{radford2021learning}. For example,  Yuan et al.~\cite{yuan2023describe} presented a method to describe facial expressions by linking image encoders and large language models, enabling the generation of textual descriptions of facial expressions that can be used for various applications. As shown in Fig.~\ref{fig:figCEprompt}, the cross-modal emotion-aware prompting (CEPrompt)~\cite{CEPrompt_zff_2024} using VLP models, incorporated emotion conception-guided visual adapter for emotion-guided visual representation, and conception-appearance tuner for optimizing cross-modal interactions, with knowledge distillation preserving pretrained knowledge, resulting in enhanced understanding of expression-related facial details.

\begin{figure}[!t]
\vspace{-0.1cm}
  \centering
   \includegraphics[width=0.9\linewidth]{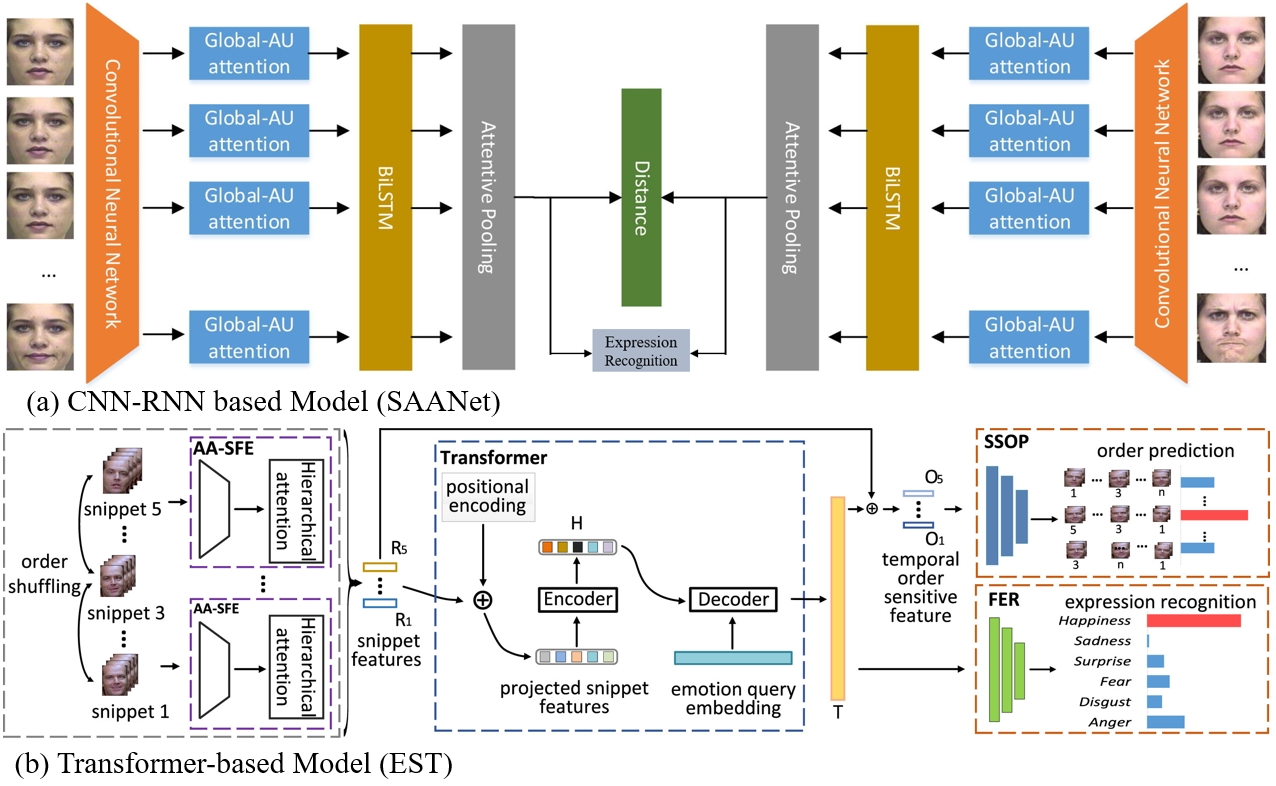}
   \vspace{-0.2cm}
   \caption{The architecture of general DFER. Figure is reproduced based on (a) CNN-RNN based model (SAANet)~\cite{liu2020saanet} and (b) the transformer-based model (EST)~\cite{liu2023expression}.}
   \label{fig:generalDFER}
   \vspace{-0.5cm}
\end{figure}

\section{Video-based Dynamic FER}
\label{sec:Dynamic}

The video-based DFER~\cite{9794419_Geometry, liu2023expression} involves analyzing facial expressions that change over time, necessitating a framework that effectively integrates spatial and temporal information. The core objective of DFER is to extract and learn the features of expression changes from video sequences or image sequences. Due to the complexity and diversity of input video or image sequences~\cite{otberdout2020dynamic}, DFER faces various task challenges. Based on different solution approaches, these challenges can be categorized into seven basic types: general DFER, sampling-based DFER, expression intensity-aware DFER, multi-modal DFER, static to dynamic FER, self-supervised DFER, and cross-modal DFER.

\subsection{General DFER}
\label{sec:Dynamic_General}

General DFER methods~\cite{sun2021multi, wang2022dpcnet, liu2023expression} primarily extract spatial-temporal features to analyze the dynamic changes in expressions. The CNN-RNN based models often combines CNNs and RNNs, while the transformer-based approach leverages deep attention mechanisms to handle more complex dynamic relationships.

\subsubsection{CNN-RNN based Models}

The early DFER approaches often utilized cascaded CNNs with RNNs to extract spatial and temporal features, such as STC-NLSTM and SAANet~\cite{yu2018spatio, liu2020saanet}. As shown in Fig.~\ref{fig:generalDFER}(a), the conjoined action-unit attention network (SAANet)~\cite{liu2020saanet} introduced a sparse self-attention mechanism for perceiving action-unit (AU) features, coupled with a twin sampling strategy and metric learning. Similarly, the multi-task global-local network ~\cite{MGLN_2020facial} integrated shared shallow, part-based, and global modules to extract spatio-temporal features from both local regions and the entire face. Chen et al. \cite{chen_9209166_2023} emphasized the exploitation of spatial-temporal and channel correlations through attention mechanisms.



\subsubsection{Transformer-based Models}

Transformer-based DFER methods excel in handling complex temporal dependencies and capturing global features by modeling the nuances and long-range relationships in facial expression sequences \cite{zhao2021former}. As shown in Fig.~\ref{fig:generalDFER}(b), Liu et al.\cite{liu2023expression} introduced the Expression Snippet Transformer (EST), which decomposes expression movements into snippets, enhancing the Transformer's capability for both intra- and inter-snippet visual modeling. Similarly, Li et al. \cite{lishutao_10095448_2023} proposed a unified spatial-temporal transformer that captures discriminative features within frames while modeling contextual relationships across frames, optimized by a compact soft maximum cross-entropy loss. Zhao et al.\cite{9794419_Geometry} developed a geometry-guided framework, combining graph convolutional networks and transformers to construct a spatial-temporal graph based on facial landmarks and local appearance, effectively representing facial expression sequences. Additionally, Poux et al.\cite{Poux9633253_2023} tackled partial facial occlusion by reconstructing the occluded regions in the optical flow domain using an auto-encoder with skip connections.

\begin{figure}[t]
\vspace{-0.1cm}
  \centering
   \includegraphics[width=0.85\linewidth]{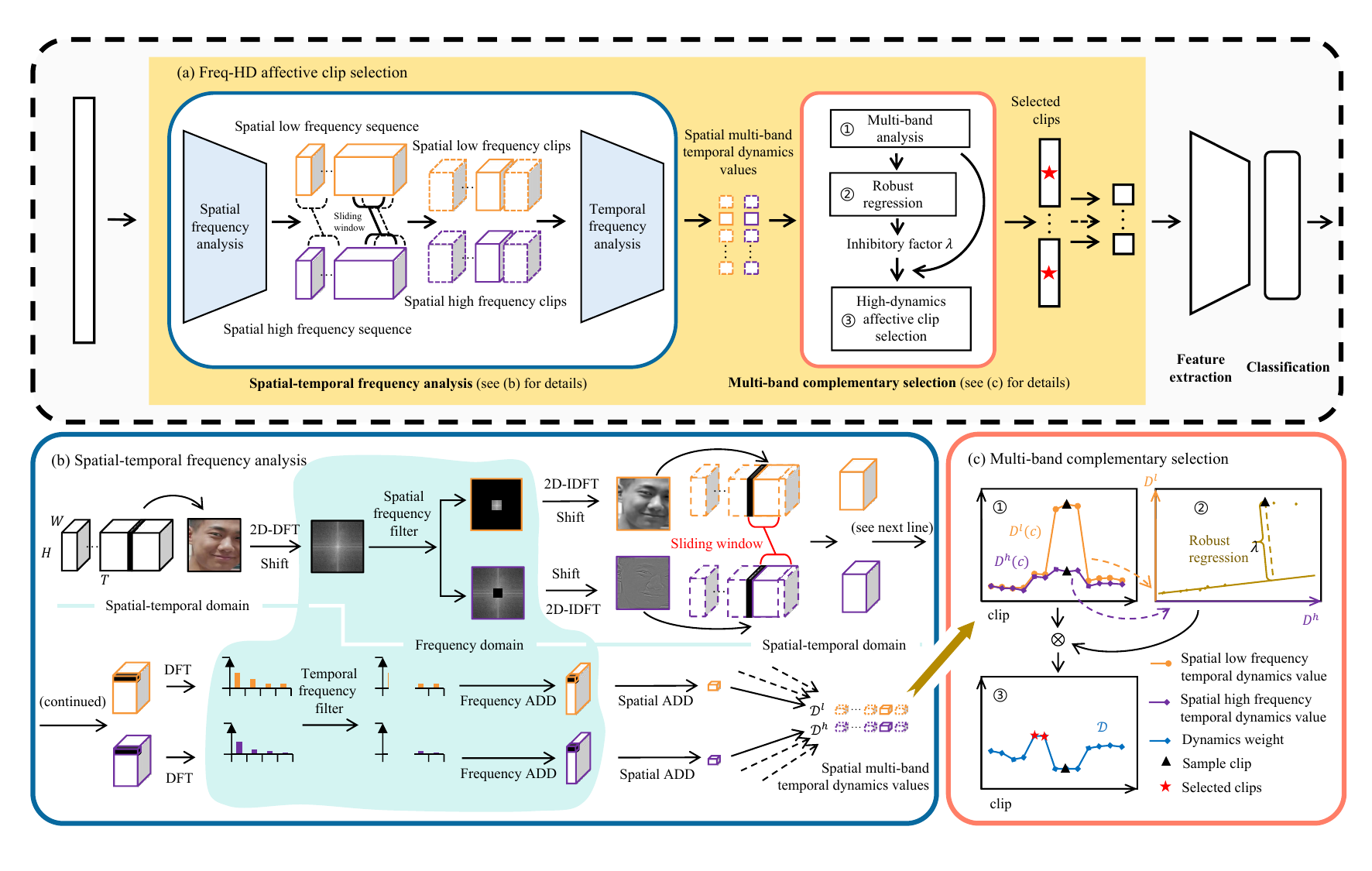}
   \vspace{-0.2cm}
   \caption{The architecture of sampling-based DFER. Figure is reproduced based on explainable sampling (Freq-HD)~\cite{tao2023freq}.}
   \label{fig:samplebased_dfer}
   \vspace{-0.2cm}
\end{figure}

\subsection{Sampling-based DFER}
\label{sec:Dynamic_Sampling}

A complete dynamic facial expression lasts about 0.5 to 4 seconds \cite{ben2021video}, typically encompassing the entire process from onset to peak and then to the end of the expression. Due to variations in capture devices and the frame rates, the sampling-based DFER aims to select expression frames, while remove interference frames and invalid frames from dynamic facial expression sequences.


\subsubsection{Random/Uniform Frame Sampling}

In Dynamic Facial Expression Recognition (DFER), two primary frame sampling methods—random and uniform sampling—are commonly employed. Random sampling~\cite{jiang2020dfew}, which involves the arbitrary selection of frames from a sequence, is valued for its simplicity and computational efficiency by mitigating over-reliance on specific frames; however, it risks overlooking key expression changes. Conversely, uniform sampling~\cite{wang2022ferv39k, zhao2021former, li2023intensity} systematically selects a predetermined number of frames, thereby ensuring comprehensive coverage of the expression sequence, which is particularly advantageous for longer videos, though it demands greater computational resources. 





\subsubsection{Explainable Frame Sampling}

Explainable frame sampling in DFER enhances conventional methods by automatically selecting emotion-rich key frames, improving model interpretability and decision-making. As shown in Fig.~\ref{fig:samplebased_dfer}, Tao et al. \cite{tao2023freq} developed Freq-HD, which utilizes Spatio-Temporal Frequency Analysis (STFA) and Multi-Band Complementary Selection (MBC) to detect significant emotional shifts, effectively distinguishing expression dynamics from irrelevant variations. Similarly, Savchenko et al. \cite{savchenko2023facial} introduced an adaptive frame rate method that adjusts sampling based on expression complexity and model confidence, optimizing frame selection for improved accuracy and efficiency. These advancements underscore the critical role of explainable frame sampling in enhancing the performance and transparency of DFER. Besides, Wang et al. \cite{wang2022dpcnet} developed a dual-path multi-excitation collaboration network incorporating space-frame and channel-time modules to learn complementary representations in manner of online frame extraction.



\begin{figure}[t]
\vspace{-0.1cm}
  \centering
   \includegraphics[width=0.85\linewidth]{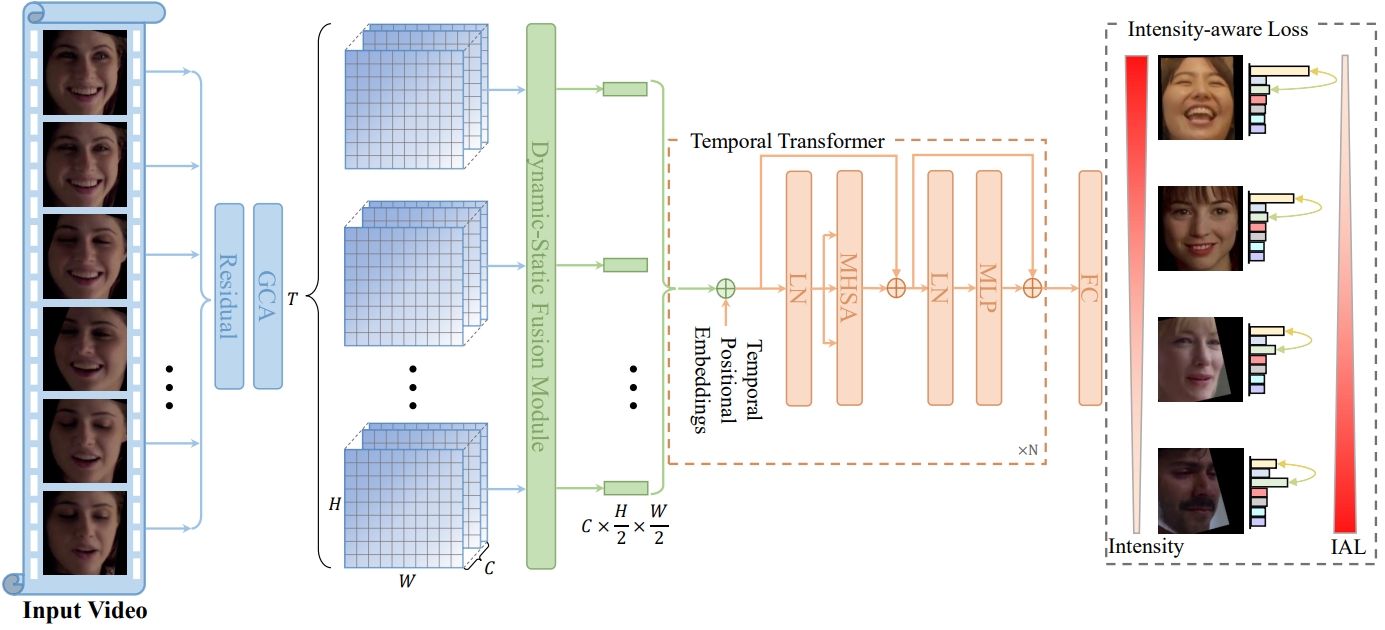}
   \vspace{-0.2cm}
   \caption{The architecture of sampling-based DFER. Figure is reproduced based on the GCA-IAL~\cite{li2023intensity}.}
   \label{fig:Intensitiy_aware_dfer}
   \vspace{-0.5cm}
\end{figure}

\subsection{Expression Intensity-aware DFER}
\label{sec:Dynamic_Expression}

Facial expressions are inherently dynamic, with intensity either gradually shifting from neutral to peak and back or abruptly transitioning from peak to neutral~\cite{otberdout2020dynamic}, making the accurate capture of these fluctuations essential for understanding expression dynamics. Early research primarily focused on modeling the temporal progression of expressions and transitions between intensity levels~\cite{zhao2016peak}, such as Zhao et al.'s~\cite{zhao2016peak} use of peak-guided deep networks (PGDN) for feature extraction and peak gradient suppression during training. Recently, Li et al.\cite{li2023intensity} developed a GCA-IAL model, including a global convolution-attention module (GCA) and a temporal transformer to learn long-distance dependencies between frames, and an expression intensity perception loss function (IAL) to discriminate low-intensity expressions as illustrated in Fig.~\ref{fig:Intensitiy_aware_dfer}. Additionally, Wang et al.\cite{Liuqs_9134869_2022} advanced the exploration of temporal expression dynamics by proposing a phase space reconstruction network to represent expression trajectories, while CEFLNet~\cite{liu2022clip} introduced a clip-based feature encoder (CFE) with cascaded self-attention for spatio-temporal feature encoding.

\subsection{Static to Dynamic FER}
\label{sec:Dynamic_s2d}

The static to dynamic FER utilized the high-performance SFER knowledge to explore appearance features and dynamic dependencies. The early work, such as Multi-channel Deep Spatial-Temporal feature Fusion neural Network (MDSTFN) \cite{sun2019deep} leverages pretrained deep CNNs for effective feature extraction and fusion in static images. Recently, Static-to-Dynamic model (S2D)~\cite{Hong2023static} utilized existing SFER knowledge and dynamic information from facial landmark-aware features to enhance the performance of DFER. Specifically, the SFER model is first built with a Vision Transformer (ViT) and Multi-View Complementary Prompters (MCPs). The temporal-modeling adapters (TMAs) are then added to the DFER model. MCPs improve facial expression features with landmark-aware data, while TMAs capture and model dynamic facial expression changes, extending the pre-trained image model to video. Similarly, an affectivity extraction network (AEN)~\cite{Lee_2023_CVPR} integrated multi-level semantic features and emotion-guided loss functions to enhance sentiment and specific emotion classification, ensuring the preservation of emotional information across video sequences.

\begin{figure}[t]
\vspace{-0.1cm}
  \centering
   \includegraphics[width=0.85\linewidth]{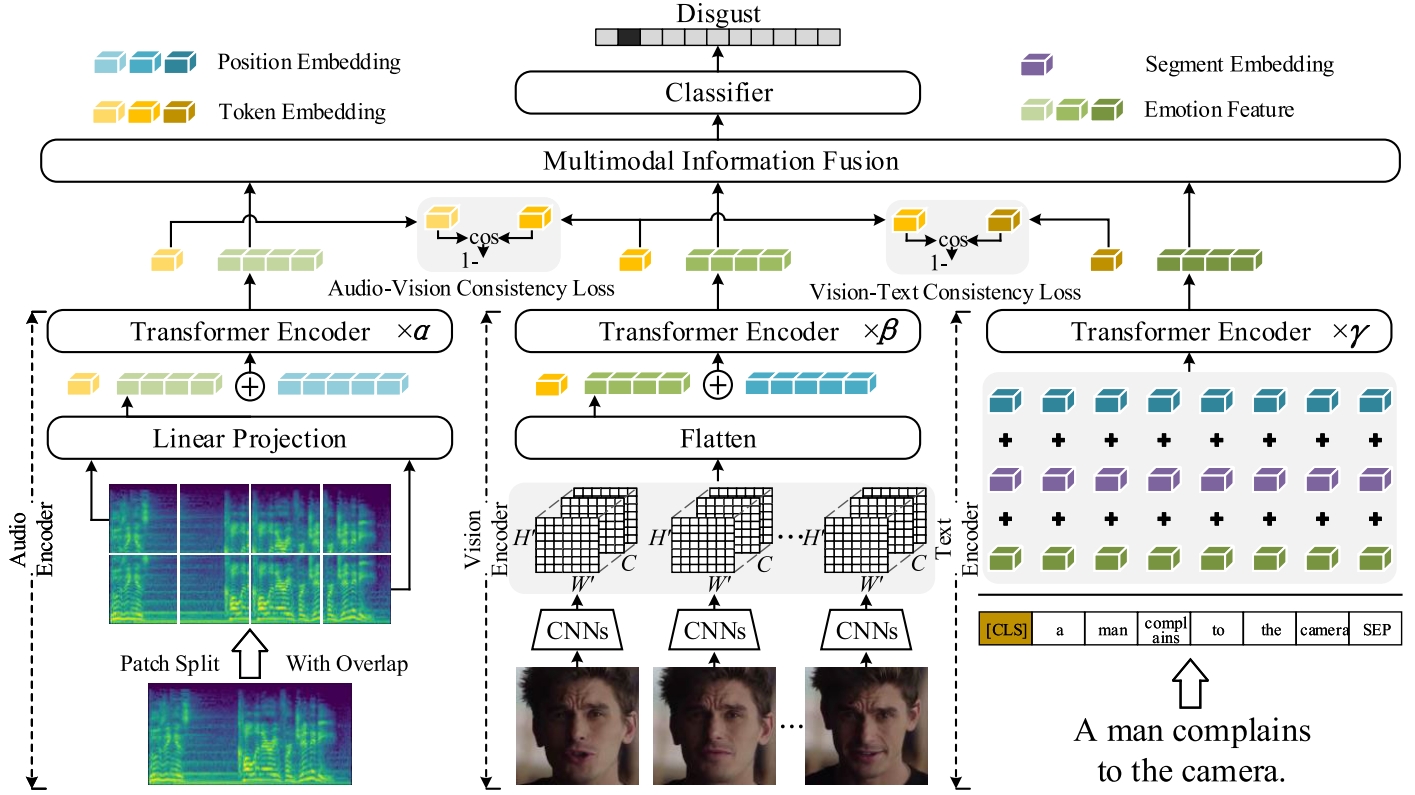}
   \vspace{-0.1cm}
   \caption{The architecture of multi-modal DFER. Figure is reproduced based on the fusion-based model (T-MEP)~\cite{10250883_dfer}.}
   \label{fig:multimodaldfer}
   \vspace{-0.5cm}
\end{figure}

\subsection{Multi-modal DFER}
\label{sec:Dynamic_multimodal}

Inspired by the affective image content analysis (AICA) \cite{Zhao_9472932} comprehending the emotional impact of images necessitates the integration of visual features with contextual cues, multi-modal DFER works \cite{lee2019context, liu2022mafw, mai2024ous} tried to leverage contexual features and fused information to capture and analyze the dynamic changes in facial expressions.





\subsubsection{Context-aware Models}
The standard DFER approach involves segmenting the face region from video or image sequences to extract expression features and classify emotions, however often overlooks crucial contextual information which is important for DFER. To exploit a joint fusion of human facial expression and context information, the context-aware emotion recognition (CAERNet) \cite{lee2019context} utilized two sub-networks to separately extract the features of face and context regions, and adaptive fusion networks to fuse such features in an adaptive fashion. Similarly, to tackle the rigid cognitive problem of DFER models which filter out environmental cues and body language, focusing only on facial information, the Overall Understanding of the Scene (OUS) \cite{mai2024ous} leveraged AudioCLIP to integrate scene and facial features.



\subsubsection{Fusion-based Models}

Fusion-based DFER integrates speech and text to enhance the accuracy and comprehensiveness by capturing auditory cues like tone and pitch, while text analysis provides emotional context. Liu et al. \cite{liu2022mafw} built a multi-modal affective dataset (MAFW), and proposed a novel Transformerbased expression snippet feature learning method to enhance learning from both facial expressions and combined emotional states over time. Besides, the model structures of spatiotemporal neural network methods (such as CNN-LSTM and C3D-LSTM)~\cite{He_2016_CVPR, tran2015learning, RNN_1997} are used to extract and fusion multi-modal information with video, audio, and text. Recently, as shown in Fig.~\ref{fig:multimodaldfer}, Zhang et al. \cite{10250883_dfer} introduced a Transformer-based Multimodal Emotional Perception (T-MEP) framework that integrates audio, image, and text sequences to bolster the robustness of expression recognition. By utilizing transformer-based encoders and a multimodal fusion module, T-MEP effectively synthesizes diverse emotional cues, resulting in enhanced performance in complex real-world environments.

\begin{figure}[t]
\vspace{-0.1cm}
  \centering
   \includegraphics[width=0.85\linewidth]{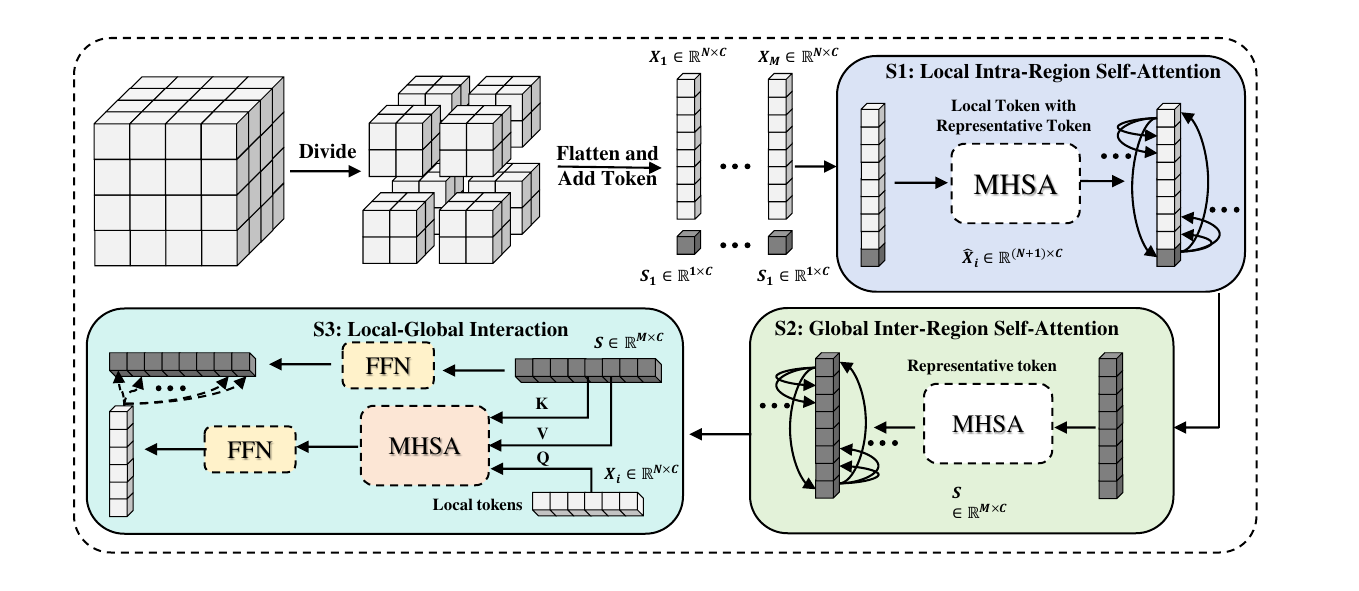}
   \vspace{-0.2cm}
   \caption{The architecture of self-supervised DFER. This is reproduced based on the MAE-DFER \cite{sun2023mae}.}
   \label{fig:supervised_DFER}
   \vspace{-0.2cm}
\end{figure}


   \begin{figure}[t]
\vspace{-0.1cm}
  \centering
   \includegraphics[width=0.85\linewidth]{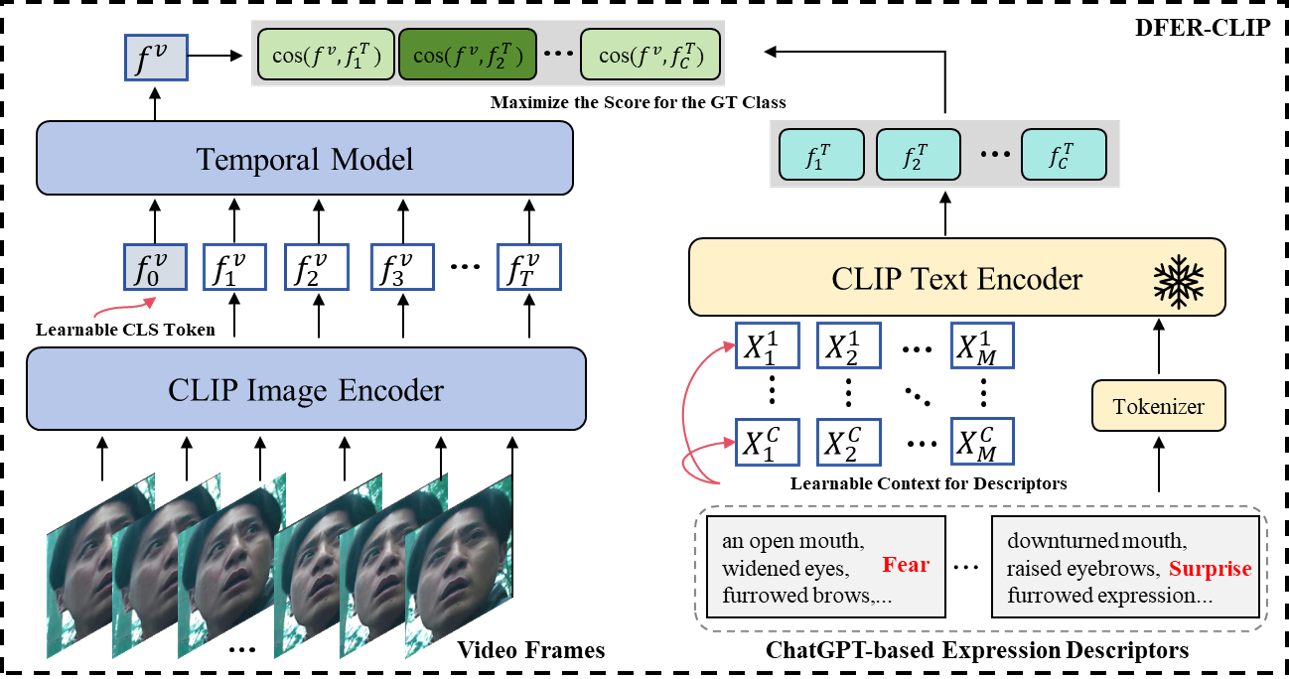}
   \vspace{-0.2cm}
   \caption{The architecture of vision-language DFER. Figure is reproduced based on DFER-CLIP~\cite{zhao2023dferclip}.}
   \label{fig:figDFER_CLI}
   \vspace{-0.5cm}
\end{figure}


\subsection{Self-supervised DFER}
\label{sec:Dynamic_selfsupervised}

The self-supervised DFER aims to learn useful representations from unlabeled video data, capturing the temporal dynamics and subtle variations in facial expressions. Specifically, Li et al.~\cite{li2020learning} proposed a twin-cycle autoencoder (TAE) to learn discriminative representations for facial actions from unlabeled videos. TAE disentangles facial actions from head motions by evaluating the quality of synthesized images, effectively capturing the subtle nuances of facial expressions. As shown in Fig.~\ref{fig:supervised_DFER}, the MAE-DFER \cite{sun2023mae} utilized large-scale unlabeled facial video data for self-supervised pre-training based on the masked autoencoders. The MAE-DFER incorporated an efficient local-global interaction Transformer (LGI-Former) as the encoder, and further integrated explicit temporal facial motion modeling alongside static appearance reconstruction.






\subsection{Visual-Language DFER}
\label{sec:Dynamic_vislan}

The visual-language DFER can extract meaningful features from facial sequences and match them with corresponding textual descriptions, enabling a more nuanced understanding of emotional expressions. As shown in Fig.~\ref{fig:figDFER_CLI}, the DFER-CLIP~\cite{zhao2023dferclip} integrated visual and textual components via CLIP-based model. The visual part employs a CLIP-based image encoder with a temporal model using Transformer encoders to extract temporal facial expression features, while the textual part uses large language models like ChatGPT to generate descriptive inputs, enhancing the accuracy of expression recognition by capturing contextual relationships. In contrast, CLIPER \cite{li2023cliper} enhanced interpretability by introducing Multiple Expression Text Descriptors (METD) to learn fine-grained expression representations. EmoCLIP \cite{foteinopoulou_emoclip_2024} further extends this approach by incorporating contextual information from the environment surrounding facial expressions, enabling zero-shot classification of emotions.


\begin{table}[!t]
    \centering
\caption{Performance (WAR) of image-based SFER and video-based DFER methods on four in-the-lab datasets.}
\renewcommand\arraystretch{1.4} 
\setlength{\tabcolsep}{0.4pt} 
\scalebox{0.8}{ 
    \begin{tabular}{
        >{\raggedright\arraybackslash}p{2.5cm} 
        >{\centering\arraybackslash}p{0.9cm} 
        >{\centering\arraybackslash}p{1.4cm} 
        >{\centering\arraybackslash}p{1.8cm} 
        >{\centering\arraybackslash}p{0.9cm} 
        >{\centering\arraybackslash}p{0.9cm} 
        >{\centering\arraybackslash}p{2.0cm}
        }
        \toprule
        \multirow{2}{*}{\textbf{Method}} & \multirow{2}{*}{\textbf{Year}} & \multirow{2}{*}{\textbf{Type}} & \multirow{2}{*}{\textbf{Backbone}} & \multicolumn{3}{c}{\textbf{Datasets}} \\ \cline{5-7}
        & & &  & \textbf{MMI} & \textbf{CK+}  & \textbf{Oulu-CASIA}\\
        \midrule
        IL-VGG~\cite{cai2018island}     & 2018  & Static  & VGG-16         & 74.68  & 91.64  & 84.58 \\
        FMPN~\cite{8965826}             & 2019  & Static  & CNNs         & 82.74  & 98.60  & -\\
        LDL-ALSG~\cite{chen2020label}   & 2020  & Static  & ResNet-50         & 70.03  & 93.08  & 63.94\\
        IE-DBN~\cite{9244224}           & 2021  & Static   & VGG-16        & -      & 96.02  & 85.21\\
        im-cGAN~\cite{SUN2023109157}    & 2023  & Static   & GAN        & -      & 98.10  & 93.34 \\
        Mul-DML~\cite{YANG2024110711}   & 2024  & Static   & ResNet-18   & 81.57  & 98.47  & -\\
        \bottomrule
        STC-NLSTM~\cite{yu2018spatio}      & 2018  & Dynamic  & 3DCNN      & 84.53  & 99.80  & 93.45 \\
        SAANet~\cite{liu2020saanet}    & 2020  & Dynamic & VGG-16        & -      & 97.38  & 82.41 \\
        MGLN~\cite{MGLN_2020facial}        & 2020  & Dynamic & VGG-16       & -      & 98.77  & 90.40\\ 
        MSDmodel~\cite{sun2021multi}       & 2021  & Dynamic & CNN       & 89.99  & 99.10  & 87.33\\ 
        DPCNet~\cite{wang2022dpcnet}   & 2022  & Dynamic & CNN       & -  & 99.70  & -\\
        STACM~\cite{chen_9209166_2023}     & 2023  & Dynamic & CNN       & 82.71  & 99.08  & 91.25\\
        \bottomrule
    \end{tabular}
}
\label{tab:in_the_lab_performance}
   \vspace{-0.3cm}
\end{table}


\section{Recent Advances of FER on Benchmark Datasets}
\label{sec:Discussion}

We have reviewed the task challenges and network models for FER with static and dynamic emotions. Below we compared the performance of the image-based SFER \textbf{(Sec. \ref{sec:Static})} and video-based DFER methods \textbf{(Sec. \ref{sec:Dynamic})} in the lab or wild scenes, and summarized their recent advances.

\subsection{Recent Advances of In-the-lab FER}

Table~\ref{tab:in_the_lab_performance} shows evaluations on four widely adopted in-the-lab datasets. Note it shows the best FER performance as FER pre-training often have different implementations. Three conclusions can be drawn from Table~\ref{tab:in_the_lab_performance}: 1) Significant progress has been made in image-based SFER and video-based DFER within laboratory environments. Due to the small scale, homogeneity, and high quality of datasets such as JAFFE, MMI, and CK+, advanced DL-based models have achieved recognition accuracies typically exceeding 95\%; 2) Since Oulu-CASIA dataset contains videos under diverse illumination settings, all the models perform much less accurately (less than 90\%) than they do on other datasets. This makes it particularly valuable for evaluating the robustness and generalization capabilities of FER models, providing a comprehensive testbed for assessing the impact of environmental variations on recognition accuracy; 3) In the widely used DFER datasets, have achieved near-perfect accuracy by effectively capturing both temporal and spatial information, with recognition rates reaching 99\% and 90\%, respectively. These results are notably higher than those obtained using datasets based on single expression frames, covering the importance of temporal dynamics in enhancing recognition performance. 

As Table~\ref{tab:in_the_lab_performance} shows, MSDmodel~\cite{sun2021multi}, DPCNet~\cite{wang2022dpcnet}, and im-cGAN~\cite{SUN2023109157} achieve state-of-the-art performance on MMI, CK+, and Oulu-CASIA, reaching 89.99\%, 99.70\%, and 93.34\%, respectively; the MSDmodel~\cite{sun2021multi} performs well consistently across three datasets. While performance on these datasets is consistently high (often greater than 90\%), the robustness and generalization of DL-based models in complex real-world scenarios remain further exploration.



\begin{table}[!t]
    \centering
\caption{Performance (WAR) of image-based SFER methods on three in-the-wild datasets.}
\renewcommand\arraystretch{1.4} 
\setlength{\tabcolsep}{0.4pt} 
\scalebox{0.8}{
    \begin{tabular}{
        >{\centering\arraybackslash}p{1.8cm} 
        >{\raggedright\arraybackslash}p{2.2cm} 
        >{\centering\arraybackslash}p{0.6cm} 
        >{\centering\arraybackslash}p{1.8cm} 
        >{\centering\arraybackslash}p{0.9cm} 
        >{\centering\arraybackslash}p{1.3cm} 
        >{\centering\arraybackslash}p{1.3cm}}
        \toprule
        \multirow{2}{*}{\makecell{\textbf{Task} \\ \textbf{Challenges}}} & \multirow{2}{*}{\textbf{Method}} & \multirow{2}{*}{\textbf{Year}} & \multirow{2}{*}{\textbf{Backbone}} & \multicolumn{3}{c}{\textbf{Datasets}} \\ \cline{5-7}
        & & & & \textbf{SFEW} & \textbf{RAF-DB} & \textbf{AffectNet} \\
        \midrule
        \multirow{10}{*}{\makecell{General \\ SFER \\ \textbf{(Sec. \ref{sec:Static_general})}}} 
        & IFSL~\cite{yan2020low}           & 2020  & VGG16        & 46.50  & 76.90   & -      \\
        & OAENet~\cite{wang2021oaenet}     & 2021  & VGG16        & -  & 86.50          & 58.70  \\
        & MA-Net \cite{zhao2021learning}   & 2021  & ResNet18 & - & 88.40          & 64.53   \\
        & D\textsuperscript{3}Net~\cite{mo2021d3net}         & 2021  & ResNet18 & 62.16 & 88.79         & -       \\
        & TransFER~\cite{xue2021transfer}   & 2021  & ResNet50 &-  & 90.91         & 66.23   \\
        & VTFF ~\cite{ma2021facial}        & 2023  & Transformer & - & 88.14       & 61.85   \\
        & HASs ~\cite{Liu10173748}        & 2023  & ResNet50    & 65.14  & 91.04       & -       \\
        & APViT ~\cite{xue2022vision}     & 2023  & Transformer & 61.92 & 91.98       & 66.91  \\
        & POSTER~\cite{Chen_10350905}    & 2023 & CNN-IR50 & - & 92.05             & 67.31   \\
        & MGR\textsuperscript{3}Net~\cite{wang2024mgr}    & 2024 & ResNet50 & - & 91.05             & 66.36   \\ 
        \midrule
        \multirow{8}{*}{\makecell{Disturbance \\ -invariant \\ SFER \\ \textbf{(Sec. \ref{sec:Static_Disturbance})}}}  \
        & PG-Unit~\cite{19_li2018occlusion}  & 2018 & VGG16     & - & 83.27       & 55.33   \\
        & IDFL~\cite{9511448_FER}           & 2021 & ResNet50   & -  & 86.96       & 59.20   \\
        & FDRL~\cite{ruan2021feature}      & 2021  & ResNet18   & 62.16 & 89.47       & -       \\
        & AMP-Net~\cite{Liu2022Ada}        & 2022 & ResNet50    & -  & 88.06       & 63.23  \\   
        & PACVT~\cite{liu2023patch}        & 2023 & ResNet18 & - & 88.21          & 60.68   \\
        & IPD-FER~\cite{jiang2022disentangling} & 2023 & ResNet18 & 58.43 & 88.89     & -       \\
        & Latent-OFER~\cite{Lee_2023_ICCV} & 2023 & ResNet18 & - & 89.60          & -       \\
        & RAC+RSL~\cite{zhang2024leave}    & 2023 & ResNet18 & - & 89.77          & 62.16   \\
        \midrule
        \multirow{5}{*}{\makecell{Uncertainty \\ -aware \\ SFER \\ \textbf{(Sec. \ref{sec:Static_Uncertainty})}}} 
        & SCN~\cite{wang2020suppressing}    & 2020 & ResNet18 & -  & 87.03         & 60.23   \\
        & DMUE~\cite{she2021dive}           & 2021 & ResNet18 & 57.12 & 88.76         & 62.84  \\
        & RUL~\cite{zhang2021relative}      & 2021 & ResNet18 & - & 88.98         & -       \\
        & EASE~\cite{Yang2022ease}          & 2022 & VGG16    & 60.12 & 89.56         & 61.82   \\
        & EAC~\cite{Deng2022learn}          & 2022 & ResNet18 & -  & 89.99         & 65.32  \\
        & LA-Net~\cite{wu2023net}           & 2023 & ResNet18 & - & 91.56         & 64.54   \\
        & LNSU-Net~\cite{zhang2024leave}    & 2024 & ResNet18 & -  & 89.77         & 65.73  \\
        \midrule
        \multirow{3}{*}{\makecell{Weak \\ -supervised \\ SFER \\ \textbf{(Sec. \ref{sec:Static_Weak})}}} 
        & Ada-CM~\cite{li2022towards}          & 2022 & ResNet18 & 52.43 & 84.42      & 57.42 \\
        & E2E-WS~\cite{Xucs_9403940}           & 2022 & ResNet18 & 54.56 & 88.89      & 60.04 \\
        & DR-FER~\cite{li2023dr}               & 2023 & ResNet50 & -  & 90.53      & 66.85  \\
        & WSCFER~\cite{liuhh_wscl}             & 2023 & IResNet  & - & 91.72      & 67.71  \\
        \midrule
        \multirow{3}{*}{\makecell{Cross-modal \\ SFER \\ \textbf{(Sec. \ref{sec:Static_Crossmodal})}}} 
        & CLEF~\cite{Zhang_2023_ICCV} & 2023 & CLIP & - & 90.09                   & 65.66     \\
        & VTA-Net~\cite{lv2024visual} & 2024 & ResNet-18 & - & 72.17              & -        \\
        & CEPrompt~\cite{CEPrompt_zff_2024} & 2024 & ViT-B/16 & - & 92.43         & 67.29    \\       
        \bottomrule
    \end{tabular}
}
\label{tab:sfer_performance}
\end{table}

\begin{table}[!t]
\caption{Performance (Accuracy) of 3D SFER methods \textbf{(Sec. \ref{sec:Static_3D})} on BU-3DE and Bosphorus datasets}
\renewcommand\arraystretch{1.4} 
\setlength{\tabcolsep}{0.4pt} 
\centering
\scalebox{0.8}{ 
    \begin{tabular}{
        >{\raggedright\arraybackslash}p{2.4cm} 
        >{\centering\arraybackslash}p{0.8cm} 
        >{\centering\arraybackslash}p{1.8cm} 
        >{\centering\arraybackslash}p{1.8cm} 
        >{\centering\arraybackslash}p{1.8cm} 
        >{\centering\arraybackslash}p{1.8cm}}
        \toprule
        \multirow{2}{*}{\textbf{Method}} & \multirow{2}{*}{\textbf{Year}} & \multirow{2}{*}{\textbf{Backbone}} & \multirow{2}{*}{\textbf{Modality}} & \multicolumn{2}{c}{\textbf{Datasets}} \\ \cline{5-6}
        & & & & \textbf{BU-3DE} 
        & \textbf{Bosphorus} \\
        \midrule
        
        JPE-GAN\cite{zhang2018joint} & 2018 & CNN &  2D/ & 81.20/- & -/- \\
        DA-CNN~\cite{zhu2019discriminative} & 2019 & ResNet50 & -/3D & -/87.69 & -/- \\


       GAN-Int~\cite{yang2021intensity} & 2021 & VGGNet16 & 2D+3D/3D & 88.47/83.20 &  -/- \\

        FFNet-M~\cite{sui2021ffnet} & 2021 & VGGNet16 & 2D+3D/3D & 89.82/87.28 & 87.65/82.86 \\

        CMANet~\cite{zhu2022cmanet} & 2022 & VGGNet16 & 2D+3D/3D & 90.24/84.03 & 89.36/81.25 \\


        DrFER~\cite{li2024drfer} & 2024 & ResNet18 & -/3D & -/89.15 & -/86.77 \\
        
        \bottomrule
    \end{tabular}
}
\label{tab:3d_sfer}
   \vspace{-0.2cm}

\end{table}

\subsection{Recent Advances of In-the-wild SFER}

Table~\ref{tab:sfer_performance} shows results on three widely adopted in-the-wild datasets. Five conclusions can be drawn from Table~\ref{tab:sfer_performance}: 1) \textbf{Significantly lower performance} (on average 20\%) in the open environment compared to the image-based SFER in the controlled laboratory environment (Table~\ref{tab:in_the_lab_performance}); 2) \textbf{Substantial variability of best performances} of FER models is across different benchmark datasets, such as 50\%-60\% in SFEW, 80\%-93\% in RAF-DB, and 55\%-67\% in AffectNet; 3) \textbf{face occlusion and pose changes} often cause the critical information loss of facial part region information, hence obtaining available facial regions and effectively extracting critical facial expressive features are the main ways to overcome disturbance. The attention-based models~\cite{9511448_FER, liu2023patch} often utilized patch or region attention CNNs to perceive occluded regions and capture salient affective interactions, however decomposition-based models~\cite{jiang2022disentangling} decompose facial expression from identity and posture, and generate discriminative facial expression features;  4) \textbf{Label and data uncertainty} mainly arises from inherent data ambiguity and subjective judgment differences among annotators. By integrating the noise label learning~\cite{Deng2022learn} and noise-insensitive loss~\cite{zhang2024leave}, uncertainty-aware SFER~\cite{Yang2022ease, Zhaogy_2023uncertain} considers these uncertainty factors when recognizing facial expressions, not only classifying the expressions but also evaluating and handling the uncertainty of each classification result to improve the accuracy and reliability of FER models; 5) \textbf{Benefit of large-scale unlabeled data and pretrained models} can improve the accuracy of SFER by leveraging the facial priors knowledge learned from high-confidence predictions to label unlabeled data~\cite{li2022towards} or visual language pre-training (VLP)~\cite{radford2021learning}.

\begin{table}[!t]
\caption{Performance (WAR) of cross-domain SFER methods \textbf{(Sec. \ref{sec:Static_Cross})} on four widely-used datasets}
\renewcommand\arraystretch{1.4} 
\setlength{\tabcolsep}{0.4pt} 
\scalebox{0.85}{ 
    \centering
    \begin{tabular}{
        >{\raggedright\arraybackslash}p{1.8cm} 
        >{\centering\arraybackslash}p{0.7cm} 
        >{\centering\arraybackslash}p{1.5cm} 
        >{\centering\arraybackslash}p{1.6cm} 
        >{\centering\arraybackslash}p{0.8cm} 
        >{\centering\arraybackslash}p{0.7cm} 
        >{\centering\arraybackslash}p{1.2cm} 
        >{\centering\arraybackslash}p{1.2cm}}
        \toprule
        \multirow{2}{*}{\textbf{Method}} & \multirow{2}{*}{\textbf{Year}} & \multirow{2}{*}{\textbf{Backbone}} & \multirow{2}{*}{\makecell{\textbf{Source} \\ \textbf{Dataset}}} & \multicolumn{4}{c}{\textbf{Target Dataset}} \\ \cline{5-8}
        \multicolumn{4}{c}{} & \textbf{JAFFE} & \textbf{CK+} & \textbf{FER-2013} & \textbf{AffectNet} \\
        \midrule
        ECAN~\cite{8993748} & 2022 & ResNet50 & RAF-DB & 57.28 & 79.77 & 56.46 & - \\
        AGRA~\cite{Chen_pami} & 2022 & ResNet50 & RAF-DB & 61.5 & 85.27 & 58.95 & - \\
        PASM~\cite{liu2021point} & 2022 & VGGNet16 & RAF-DB & - & 79.65 & 54.78 & - \\
        CWCST~\cite{li2023cross} & 2023 & VGGNet16 & RAF-DB2.0 & 69.01 & 89.64 & 57.44 & 52.66 \\
        DMSRL~\cite{9676449} & 2023 & VGGNet16 & RAF-DB2.0 & 69.48 & 91.26 & 56.16 & 50.94 \\
        CSRL~\cite{9676449} & 2023 & ResNet18 & RAF-DB & 66.67 & 88.37 & 55.53 & - \\
        \bottomrule
    \end{tabular}
}
\label{tab:cross_domain_sfer_performance}
   \vspace{-0.5cm}

\end{table}

\begin{table*}[t]
 \centering
\caption{Performance (WAR/UAR) of video-based DFER methods on four widely-used datasets. TI: Time Interpolation; DS: Dynamic Sampling; GWS: Group-weighted Sampling. $*$: Tunable Param (M)}
\renewcommand\arraystretch{1.4} 
\setlength{\tabcolsep}{0.4pt} 
\scalebox{0.8}{ 
    \begin{tabular}{
        >{\centering\arraybackslash}p{2.6cm} 
        >{\raggedright\arraybackslash}p{2.6cm} 
        >{\centering\arraybackslash}p{1.0cm} 
        >{\centering\arraybackslash}p{2.2cm} 
        >{\centering\arraybackslash}p{2.2cm} 
        >{\centering\arraybackslash}p{2.2cm} 
        >{\centering\arraybackslash}p{2.0cm} 
        >{\centering\arraybackslash}p{2.0cm} 
        >{\centering\arraybackslash}p{2.1cm} 
        >{\centering\arraybackslash}p{2.0cm}}
        \toprule
        \multirow{2}{*}{\makecell{\textbf{Task} \\ \textbf{Challenges}}} & \multirow{2}{*}{\textbf{Method}} & \multirow{2}{*}{\textbf{Year}} & \multirow{2}{*}{\makecell{\textbf{Sample} \\ \textbf{Strategies}}}   & \multirow{2}{*}{\makecell{\textbf{Backbone}}} & \multirow{2}{*}{\makecell{\textbf{Comlexity} \\ \textbf{(GFLOPs)}}}  & \multicolumn{4}{c}{\textbf{Datasets (WAR/UAR)}} \\ \cline{7-10}
        \multicolumn{6}{c}{} & \textbf{AFEW} & \textbf{DFEW} & {\makecell{\textbf{FERV39k}}} & \textbf{MAFW} \\
        \midrule

        \multirowcell{8}{General DFER \\ \textbf{(Sec. \ref{sec:Dynamic_General})}}
        & TFEN~\cite{teng2021typical}      & 2021  & TI & ResNet-18   &  -  & -   &  56.60/45.57  & - &  -  \\
        & FormerDFER~\cite{zhao2021former}  & 2021  & DS & Transformer & 9.1G & 50.92/47.42 & 65.70/53.69 & - & 43.27/31.16 \\
        & EST~\cite{liu2023expression}      & 2023  & DS & ResNet-18   & N/A  & 54.26/49.57 & 65.85/53.94 & - & - \\
        & LOGO-Former~\cite{lishutao_10095448_2023} & 2023  & DS & ResNet-18   & 10.27G  & - & 66.98/54.21 & 48.13/38.22 & - \\
        & MSCM~\cite{li2023multi}           & 2023  & DS & ResNet-18   & 8.11G & 56.40/52.30 & 70.16/58.49 & - & - \\       
        & SFT~\cite{huang2024dynamic}       & 2024  & DS & ResNet-18  & 17.52G & 55.00/50.14 & - & 47.80/35.16 & 47.44/33.39 \\
        & CDGT~\cite{chen2024cdgt}          & 2024  & DS & Transformer & 8.3G  & 55.68/51.57 & 70.07/59.16 & 50.80/41.34 & - \\
        & LSGTNet~\cite{wang2024joint}      & 2024  & DS & ResNet-18   &  -     & - & 72.34/61.33 & 51.31/41.30 & - \\
        \midrule
        
        \multirowcell{5}{Sampling-based \\ DFER \\ \textbf{(Sec. \ref{sec:Dynamic_Sampling})}}
        & EC-STFL~\cite{jiang2020dfew}    & 2020   & TI & ResNet-18 & 8.32G  & 53.26/-     & 54.72/43.60  & -  & - \\
        & DPCNet~\cite{wang2022dpcnet}    & 2022   & GWS & ResNet-50 & 9.52G  & 51.67/47.86 & 66.32/57.11  & -  & - \\
        & FreqHD~\cite{tao2023freq}     & 2023     & FreqHD & ResNet-18 & -       & - & 54.98/44.24 & 43.93/32.24 & - \\
        & M3DFEL~\cite{wang2023rethinking} & 2023  & DS & R3D18     & 1.66G  & - & 69.25/56.10 & 47.67/35.94 & - \\
        \midrule

        \multirowcell{3}{Expression \\ Intensity-aware \\ DFER \\ \textbf{(Sec. \ref{sec:Dynamic_Expression})}}
         & CEFL-Net~\cite{liu2022clip}  & 2022 & Clip-based   & ResNet-18   &  -     & 53.98/-     &  65.35/-    & -  & - \\
         & NR-DFERnet~\cite{li2022nr}   & 2023  & DS  & ResNet-18   & 6.33G & 53.54/48.37 & 68.19/54.21 & -  & - \\
         & GCA+IAL~\cite{li2023intensity} & 2023 & DS & ResNet-18   & 9.63G & -           & 69.24/55.71 & 48.54/35.82 & - \\
        \midrule

        \multirowcell{2}{Static to Dynamic \\ FER \\ \textbf{(Sec. \ref{sec:Dynamic_s2d})}}
        & S2D~\cite{Hong2023static}       & 2023  & DS & ViT-B/16   & - & - & 76.03/61.82 & 52.56/41.28 & 57.37/41.86 \\
        & (AEN)~\cite{Lee_2023_CVPR}       & 2023  & DS & Transformer   & - & 54.64/50.88 & 69.37/56.66 & 47.88/38.18 &- \\
   
        \midrule

        \multirowcell{4}{Multi-modal \\ DFER  \\ \textbf{(Sec. \ref{sec:Dynamic_multimodal})}}
        & T-ESFL~\cite{liu2022mafw}  & 2022 & DS & Transformer  &  -    & - & - & - & 48.18/33.28 \\
        & T-MEP~\cite{zhang2023transformer}  & 2023 & DS & -   & 6G   & 52.96/50.22 & 68.85/57.16 & - & 52.85/39.37 \\
        & OUS~\cite{mai2024ous}             & 2024 & DS & CLIP    & - & 52.96/50.22 & 68.85/57.16 & - & 52.85/39.37 \\
        & MMA-DFER~\cite{chumachenko2024mma} & 2024 & DS & Transformer   & - & - & 77.51/67.01 & - & 58.52/44.11 \\ 
        \midrule

       \multirowcell{2}{Self-supervised  \\ DFER  \\ \textbf{(Sec. \ref{sec:Dynamic_selfsupervised})}} 
        & MAE-DFER~\cite{sun2023mae}  & 2023 & DS & ResNet-18   & 50G  & - & 74.43/63.41 & 52.07/43.12 & 54.31/41.62 \\
        & HiCMAE~\cite{sun2024hicmae} & 2024 & DS & ResNet-18   & 32G  & - & 73.10/61.92 & - & 54.84/42.10 \\
        \midrule

      \multirowcell{5}{Visual-Language \\ DFER  \\ \textbf{(Sec. \ref{sec:Dynamic_vislan})}} 
        & CLIPER~\cite{li2023cliper}        & 2023  & DS  & CLIP-ViT-B/16  & 88M\textsuperscript{$*$}    & 56.43/52.00 & 70.84/57.56 & 51.34/41.23 & - \\
        & DFER-CLIP~\cite{zhao2023dferclip} & 2023  & DS  & CLIP-ViT-B/32  & 92G & - & 71.25/59.61 & 51.65/41.27 & 52.55/39.89 \\
        & EmoCLIP~\cite{foteinopoulou_emoclip_2024}  & 2024  & DS & CLIP-ViT-B/32  & - & - & 62.12/58.04 & 36.18/31.41 & 41.46/34.24  \\
        & A\textsuperscript{3}lign-DFER~\cite{tao20243}      & 2024  & DS & CLIP-ViT-L/14 & -  & - & 74.20/64.09  & 51.77/41.87  & 53.22/42.07 \\
        & UMBEnet~\cite{mai2024all}         & 2024   & DS & CLIP    & -  & - &73.93/64.55  &  52.10/44.01 & 57.25/46.92   \\
        & FineCLIPER~\cite{chen2024finecliper}  & 2024 & DS & CLIP-ViT-B/16   & 20M\textsuperscript{$*$} & - & 76.21/65.98 & 53.98/45.22 & 56.91/45.01 \\
        \bottomrule
    \end{tabular}
}
\label{tab:performance_dfer}
   \vspace{-0.5cm}

\end{table*}

As Table~\ref{tab:sfer_performance} shows, HASs~\cite{Liu10173748}, CEprompt \cite{CEPrompt_zff_2024}, and (WSCFER)~\cite{liuhh_wscl} achieve state-of-the-art performance on SFEW, RAF-DB, and AffectNet, reaching 65.14\%, 92.43\%, and 67.71\%, respectively; the CEprompt \cite{CEPrompt_zff_2024} performs well consistently across RAF-DB and AffectNet datasets. Besides, Table~\ref{tab:3d_sfer}, Table~\ref{tab:cross_domain_sfer_performance} show the performance ‌under the framework of‌ the 3D FER and cross-domain FER, respectively. 3D SFER utilized GAN-based learning~\cite{zhu2023var} and multi-view learning~\cite{vo20193d} to generate synthetic facial expression data with different changes and utilize multi-view images during training. Cross-domain inconsistency poses a significant challenge to the generalization of FER models, as data from controlled laboratory environments differ markedly from those in real-world applications. To address this issue, domain adaptation techniques (such as transfer learning and adversarial learning)~\cite{Zheng_10137555, Chen_pami} are employed to align data from the source and target domains, thereby reducing inter-domain differences and enhancing model robustness.

\subsection{Recent Advances of In-the-wild DFER}

Table~\ref{tab:performance_dfer} shows results on 4 widely adopted in-the-wild DFER datasets. In the past three years, significant progress of DFER has been promoted especially after the release of data sets such as DFEW~\cite{jiang2020dfew}, FERV39k~\cite{wang2022ferv39k} and MAFW~\cite{liu2022mafw}, which provide rich diversity and challenging data, covering a wider range of real-life scenarios. Five conclusions can be drawn from Table~\ref{tab:performance_dfer}: 1) \textbf{Markedly reduced performance}  (on average 30\%)  is observed in open environments compared to controlled laboratory settings (Table~\ref{tab:in_the_lab_performance}) for video-based DFER, highlighting the significant challenges of adapting to real-world conditions; 2) \textbf{Remarkable differences} of WAR/UAR performances of DFER models is across four benchmark datasets, such as 50\%/47\%$\sim$56\%/52\% in AFEW, 56\%/46\%$\sim$76\%/66\% in DFEW, 44\%/32\%$\sim$54\%/45\% in FERV39k, and 43\%/31\%$\sim$58\%/46\% in MAFW. Note the lowest performances appears in the FERV39k dataset due to the large-scale and multi-scene attributes in various real-life scenarios; 3) \textbf{Key Frame Extraction} crucial to the performance of the DFER, including selecting key frames by detecting changes in facial movements in the video~\cite{savchenko2023facial}; and extracting key frames based on changes in facial action units~\cite{tao2023freq}; 4) \textbf{Capturing expression intensity fluctuations} is pivotal for understanding the dynamic nature of expressions and enhancing the accuracy of DFER systems due to the inherently dynamic characteristic of facial expression intensities  varying over time. Since the intensity often follows two patterns: a gradual shift from neutral to peak intensity and back, or an abrupt transition from peak to neutral, PGDN~\cite{zhao2016peak} and GCA-IAL~\cite{li2023intensity} extracted features related to expression evolution and learn long-distance dependencies between frames, respectively. 5) \textbf{Leveraging the multi-modal information, large-scale unlabeled data, or pretrained models} significantly enhances DFER accuracy by utilizing facial priors acquired from the fusion of contextual features~\cite{10250883_dfer}, the masked autoencoders pretraining on large-scale unlabeled facial video data~\cite{sun2023mae}, or visual language pre-training models (CLIP)~\cite{radford2021learning}. As Table~\ref{tab:performance_dfer} shows, CLIPER~\cite{li2023cliper}, MMA-DFER~\cite{chumachenko2024mma}, FineCLIPER~\cite{chen2024finecliper}, and UMBEnet~\cite{mai2024all} achieve the best performance (the average accuracy of WAR and UAR) on AFEW, DFEW, FERV39k, and MAFW reaching 52.22\%, 72.3\%, 49.6\%, and 52.09\%, respectively; the UMBEnet~\cite{mai2024all} performs well consistently across three large-scale datasets.

\section{Applications and Ethical Issues of FER}
\label{sec:Applications}

In this section, we point out some of the applications and ethical issues of FER, which further promote technological innovation and protect individual rights and interests.

\subsection{Applications of FER}

\subsubsection{Health and Psychological Counseling}

The FER plays a pivotal role in monitoring emotional changes by analyzing users' facial expressions in real-time, providing timely psychological advice and alerts~\cite{hassan2019automatic, bisogni2022impact}. This technology is increasingly integrated into smartwatches and mobile applications, which continuously monitor emotional states and offer psychological adjustments. These devices can detect signs of depression or stress, prompting users to manage their emotions and, when necessary, connect with professional counselors~\cite{10449383_Depressed}.

In mental health monitoring, mobile apps equipped with FER capabilities offer emotion tracking and analysis, helping users understand and manage their emotional states more effectively~\cite{10449383_Depressed}. When abnormal emotions are detected, these apps can suggest relaxation techniques or direct users to seek professional help~\cite{LiuHh9416790}. In psychotherapy, particularly cognitive behavioral therapy (CBT), FER enables therapists to monitor patients' emotional reactions in real-time, enhancing their understanding of patients' internal states and allowing for personalized treatment adjustments~\cite{10449383_Depressed}. FER also aids in diagnosing psychological and neurological disorders, such as early detection of depression and Parkinson's disease, through the analysis of facial expressions and remote photoplethysmography~\cite{10023947_dr, Wang_10080949}. In special populations and scenarios, FER is valuable in understanding children's emotional states, particularly in addressing emotional disorders and behavioral issues~\cite{Zheng_10246378}. It is also applied in assessing animal emotions, such as evaluating pain levels in horses~\cite{9780602_animal_pain}, and in intensive care units, where FER can assess patient pain levels even with partial facial occlusion by analyzing facial AUs~\cite{10327783_Pain}.

\subsubsection{Personalized Education}

Monitoring students' emotional states in the classroom allows teachers to adjust their teaching methods in real time, thereby enhancing educational effectiveness~\cite{li2025application}. For instance, by analyzing students' facial expressions (such as confusion, boredom, or interest) in classroom or online education platforms, FER-enabled systems can dynamically adjust the difficulty of learning content based on students' emotional feedback or provide more detailed explanations and further materials~\cite{savchenko2022classifying}. Such a system can also facilitate timely adjustments to teaching strategies, such as adding interactive sessions and altering the teaching pace in response to emotional changes of students. Additionally, if the system identifies signs of depression or disengagement, it can prompt the teacher to offer personalized tutoring or encouraging feedback~\cite{9613750_class}.

\subsubsection{Human-Computer Interaction}

FER technology holds significant potential in enhancing human-computer interaction and robotics by making interactions more natural and personalizing emotional feedback to improve user experience~\cite{jiang_10018501}. By integrating FER, social robots can recognize users' emotional states and adjust their conversational content and emotional expressions accordingly, offering assistance when users appear confused or sharing joyful topics when users are happy. This technology also enables virtual assistants to better perceive and respond to users' emotions, providing more personalized and contextually appropriate services. Additionally, FER can drive emotion-sensitive user interfaces \cite{braun2021affective} that adapt dynamically to users' emotional responses. 

\subsection{Ethical Issues}

FER technologies offer vast applications but raise concerns regarding privacy, ethics, and security \cite{devillers2023ethical}. To ensure its responsible and sustainable development, interdisciplinary collaboration across psychology, ethics, and biology is essential \cite{wang2023unlocking}. Prolonged monitoring through FER systems can cause discomfort, anxiety, and stress, potentially leading to mental health issues and eroding trust, especially in public and work environments. Therefore, incorporating public opinion into the development process is crucial to align the technology with societal moral standards and public interest. Ethically, the research, development, and deployment of FER must be guided by a clear framework that prioritizes transparency, informed consent, and fairness. The decision-making processes of FER algorithms should be transparent and their outcomes explainable, ensuring equity across different races, genders, and age groups to prevent bias and discrimination. Given that facial expression data is a form of biometric data, it requires stringent protection against leakage and misuse, adhering strictly to ethical principles to avoid infringing on individual rights. Addressing these issues ensures that the advancement of FER technology remains aligned with social progress, balancing innovation with ethical responsibility \cite{toisoul2021estimation}.

\section{Development Trends}
\label{sec:Development}

 \textbf{Facial Action Units (AUs) assisted FER} is able to detect subtle differences that other models might overlook by emphasizing individual muscle movements \cite{liyong_2023compound}, providing a detailed and objective analysis of facial expressions and improving understanding and reliability. Defined in the Facial Action Coding System (FACS) \cite{ekman1978facial}, AUs correspond to specific muscle movements, such as raising eyebrows or wrinkling the nose. Combining different AUs allows for detailed descriptions of facial expressions, improving the accuracy, robustness, and cultural adaptability of FER models. The adaptability extends to different cultural contexts, as facial expressions and their associated muscle movements are consistent across cultures \cite{li2023contrastive}. Besides, detailed AUs \cite{snoek2023testing} enhance the interpretability of FER models, allowing researchers to understand the influence of specific facial movements on emotion recognition. This leads to models that not only accurately predict emotions but also provide clear explanations, fostering transparency and trust.

\textbf{Zero-shot FER} aims to identify emotions that the model has not encountered during training~\cite{Yang2019zero}, offering a solution when it is impractical to collect and annotate data for every possible expression. Traditional FER models, which rely heavily on large, manually labeled datasets, struggle to predict new emotion categories beyond their training data. This limitation undermines their effectiveness in real-world, dynamic scenarios where humans can express thousands of emotions~\cite{cowen2021sixteen}. The visual language models~\cite{radford2021learning} can learn robust visual features and integrate them with natural language, enabling superior zero-shot recognition guided by semantic knowledge. Leveraging pre-trained visual language models like EmoCLIP \cite{foteinopoulou_emoclip_2024}, the zero-shot FER systems use unified semantic feature learning to obtain visual and linguistic representations that generalize to unseen emotion classes, which can recognize and respond to a broader range of emotional expressions in diverse contexts.

\textbf{Multi-modal Emotion Recognition} systems aim integrating  multiple channels, including facial expressions, vocal tone, gestures, posture, and physiological signals \cite{yang_multimodal_2024}, to enhance accuracy and robustness, mirroring the human ability to perceive emotions using multiple cues. It offers a more comprehensive understanding of emotional states by capturing the full spectrum of human emotions. Single-modality systems, like those focused solely on facial expressions, can miss critical information from other channels; for instance, a smile combined with a shaky voice might indicate nervousness rather than happiness. Multi-modal systems can disambiguate such signals and provide a more nuanced understanding~\cite{Yang2023weakly}. Multimodal Large Language Models (MLLMs) \cite{bai2024survey} have introduced new possibilities for emotion recognition by aligning, pre-training, and fine-tuning multiple modalities, enabling them to understand emotions and perform zero-shot emotion recognition, demonstrating significant potential in this field.



\textbf{Embodied FER} system is essential in modern human-computer interaction~\cite{rawal2022facial}, integrating FER models with interactive technologies to achieve real-time detection and response to human emotions. These systems utilize computer vision and representation learning to analyze facial expressions, language, voice, and posture, significantly enhancing user experience and engagement. Compared to traditional camera-based FER, Embodied FER systems~\cite{lee2024encoding} face the challenge of managing dynamic, multi-perspective views and adapting to environmental variations such as lighting changes, occlusions, and motion blur in complex settings. Additionally, the need for real-time, contextually appropriate feedback during close interactions demands greater robustness, adaptability, and computational efficiency. Future research will focus on improving system performance across diverse facial morphologies and environments, and advancing the integration of multimodal methods (e.g., fusing facial expression with voice and body language) to further develop embodied FER.

\textbf{Embodied Facial Expression Generation} is crucial for enabling robots, particularly humanoid robots, to engage with humans in a direct and compelling manner by accurately mimicking facial expressions~\cite{lee2024encoding}. It can be categorized into two primary forms: AIGC-based expression generation and physical embodiment through motor-driven mechanisms. AIGC-based facial expression generation~\cite{chen2024echomimic, xu2023high} utilizes generative models, which deeply learn from vast datasets, to automatically create virtual facial expressions, which allows for a wide range of emotional expressions, contributing to more vivid and controllable interactions in dynamic environments. Physical embodiment using motor-driven mechanisms involves the movement of components such as eyes, mouth, and neck to produce facial expressions, enhancing the realism of interactions through tangible physical presence~\cite{liu2024unlocking, hu2024human}. Future research will focus on addressing the challenges associated with these methods by advancing the realism and cultural sensitivity of AIGC-generated expressions and enhancing the hardware capabilities of motor-driven systems to support more expressive and responsive facial movements.

\section{Conclusion}
\label{sec:Conclusion}

Facial expression recognition (FER) has gained significant attention within the AI community, with promising applications in human-machine collaboration and embodied intelligence. This survey extensively reviews FER works from several perspectives, including background, datasets, generic workflow, challenge-oriented taxonomy of state-of-the-art methods, recent advances, applications, ethical concerns, and emerging trends. We systematically compare and summarize FER datasets, task challenges, methods, and performance evaluations through tables and figures, providing a clear overview of the latest advancements in FER. This comprehensive analysis greatly benefits researchers from various disciplines by enabling them to swiftly understand the challenges and progress in the field, thereby fostering collaboration toward the development of general FER.

\section{Acknowledgments}
\label{sec:Acknowledgments}
This work was supported by National Natural Science Foundation of China (No.62406075), National Key Research and Development Program of China (2023YFC3604802), and by China Postdoctoral Science Foundation under Grant (2023M730647, 2023TQ0075).

\bibliographystyle{IEEEtran}
\bibliography{IEEEabrv,egbib}

%



\vspace{-2cm}

\end{document}